\documentclass[journal]{IEEEtran}

\usepackage{color}
\usepackage{caption}
\usepackage{amsmath}
\usepackage{multirow}
\usepackage{multicol}
\usepackage{subcaption}
\usepackage{graphicx}
\usepackage{epstopdf}
\graphicspath{{figure/}}

\begin{document}

\title{A Direct Slip Ratio Estimation Method based on an Intelligent Tire and Machine Learning}

\author{Nan~Xu,
Zepeng~Tang,
Hassan Askari,
Jianfeng Zhou and Amir Khajepour

\thanks{This work has been submitted to the MSSP for possible publication. Copyright may be transferred without notice, after which this version may no longer be accessible. This research is supported by the National Natural Science Foundation of China (Grant Nos. 51875236 and 61790561), China Automobile Industry Innovation and Development Joint Fund under Grant U1864206.}
\thanks{N. Xu, Z. Tang and J. Zhou are with the State Key Laboratory of Automotive Simulation and Control, Jilin University, Changchun, Jilin, 130025, China, and N. Xu is also with the Department of Mechanical and Mechatronics Engineering, University of Waterloo, ON. N2L3G1, Canada, e-mail: (xunan@jlu.edu.cn, tangzp19@mails.jlu.edu.cn, and zhoujf20@mails.jlu.edu.cn).}
\thanks{H. Askari and A. Khajepour are at the Department of Mechanical and Mechatronics Engineering, University of Waterloo, ON. N2L3G1, Canada e-mail: ( h2askari@uwaterloo.ca, and a.khajepour@uwaterloo.ca) }}

\markboth{Journal of \LaTeX\ Class Files,~Vol.~14, No.~8, August~2015}%
{Xu \MakeLowercase{\textit{et al.}}: A Direct Slip Ratio Estimation Method based on an Intelligent Tire and Machine Learning}

\maketitle

\begin{abstract}
Accurate estimation of the tire slip ratio is critical for vehicle safety, as it is necessary for vehicle control purposes. In this paper, an intelligent tire system is presented to develop a novel slip ratio estimation model using machine learning algorithms. The accelerations, generated by a triaxial accelerometer installed onto the inner liner of the tire, are varied when the tire rotates to update the contact patch. Meanwhile, the slip ratio reference value can be measured by the MTS Flat-Trac tire test platform. Then, by analyzing the variation between the accelerations and slip ratio, highly useful features are discovered, which are especially promising for assessing vertical acceleration. For these features, machine learning (ML) algorithms are trained to build the slip ratio estimation model, in which the ML algorithms include artificial neural networks (ANNs), gradient boosting machines (GBMs), random forests (RFs), and support vector machines (SVMs). Finally, the estimated NRMS errors are evaluated using 10-fold cross-validation (CV). The proposed estimation model is able to estimate the slip ratio continuously and stably using only the acceleration from the intelligent tire system, and the estimated slip ratio range can reach 30\%. The estimation results have high robustness to vehicle velocity and load, where the best NRMS errors can reach 4.88\%. In summary, the present study with the fusion of an intelligent tire system and machine learning paves the way for the accurate estimation of the tire slip ratio under different driving conditions, which create new opportunities for autonomous vehicles, intelligent tires, and tire slip ratio estimation.
\end{abstract}


\IEEEpeerreviewmaketitle

\section{Introduction}

\begin{figure}[!h]
    \centering
    \includegraphics[width=3in]{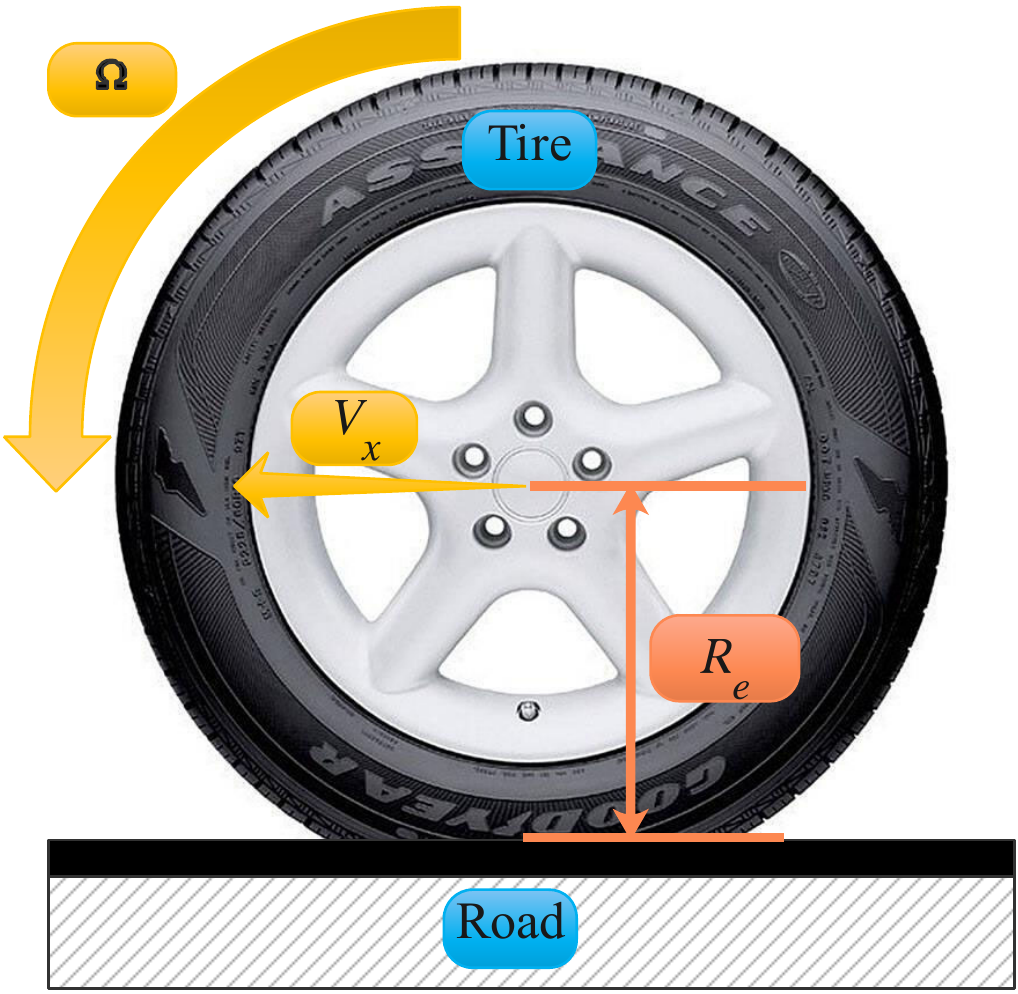}
\caption{Schematic representation of a tire interacting with road.}
    \label{fig:tire model}
\end{figure}

\IEEEPARstart{A}{s} autonomous vehicles are becoming more popular, and the need to develop intelligent sensing systems, specifically intelligent tires, has become more pronounced \cite{lee2017intelligent,askari2019tire}. Intelligent tires can provide vital information about tire dynamics, which are necessary for designing advanced vehicle controls. Among the different tire parameters, the slip ratio plays an important role in vehicle dynamics \cite{heidfeld2020ukf}. Having an excessive slip ratio reduces the braking capability. In addition, it can affect the steering ability if the front and rear wheels lose lateral adhesion. Accordingly, it is necessary to effectively control the slip ratio. For example, the anti-lock braking system (ABS) \cite{pretagostini2020survey,rafatnia2021adaptive}, electronic stability control (ESC) \cite{liebemann2004safety} and traction control system (TCS) \cite{park1999wheel} innovations that have been developed are used to ensure that the tires have optimal adhesion by limiting the tire slip ratio to an effective range. It can be derived from Eq. (\ref{eqn:01}) that the slip ratio ($\kappa$), as observed in Fig. \ref{fig:tire model}, mainly reflects the rate of the difference between the longitudinal vehicle velocity ($V_x$) and the wheel velocity ($V_w$), i.e., the level of slip of the tire relative to the road.

\begin{align}
\kappa  = \frac{V_{w}-V_{x} }{ V_{x} }  = \frac{\Omega R_{e}-V_{x}  }{ V_{x} },
\label{eqn:01}
\end{align}
where $V_w$ and $V_x$ are the wheel velocity and longitudinal velocity of the vehicle, respectively. $\Omega$ is the wheel angular velocity, and $R_e$ is the tire effective radius.

Several studies have been performed to estimate the tire slip ratio. Among the prevailing methods is using the vehicle velocity, which mainly relies on the accurate estimation of longitudinal velocity, and the main challenge is to find a potent estimation technique for this purpose. Although vehicle velocity estimation methods are a traditional topic, there are still some problems troubling the whole industry. For example, kinematic-based algorithms for vehicle velocity have been proposed by several researchers. The acceleration and yaw rate from the inertial measurement unit (IMU) are typically used to establish the vehicle velocity observer, and the methodology used includes the Kalman filter (KF) \cite{rezaeian2016simultaneous} and nonlinear observer \cite{imsland2006vehicle}. However, the problems in this method, such as the errors caused by acceleration bias and noise, are not negligible. Therefore, researchers have proposed the use of measurement signals from additional sensors to correct the kinematic-based estimated vehicle velocity. Bevly et al. \cite{bevly2004global} used accurate GPS velocity to reduce errors inherent in kinematic-based estimation. Yoon et al. \cite{yoon2013cost} developed an extended Kalman filter (EKF)-based vehicle estimation using two low-cost GPS receivers to compensate for the estimated vehicle velocity. However, low-cost GPS has low update frequency, while high-precision and high-frequency GPS has excessive cost. Moreover, such measurement devices are subject to data loss in certain environments, such as urban buildings, tunnels or tall trees \cite{rafatnia2018move}. Lv et al. \cite{lv2019vehicle} came up with a new way to correct the integral errors of the estimated vehicle velocity on a long timescale by two reliable driving empirical judgments while correcting the error on a short timescale using the estimated velocity between the adjacent strong empirical correction instants. Although the vehicle velocity is compensated, the acceleration used interacts with gravity and body posture, and under certain bank angle and road grade scenarios, additional errors will occur. Rezaeian et al. \cite{rezaeian2016simultaneous} used roll rate and pitch rate measured by a 6-axis IMU to compensate for the estimated velocity errors due to body posture changes, but 6-axis IMUs are not common in commercial vehicles.

Another major prevailing approach uses a dynamics-based model to estimate the vehicle velocity. Zhao et al. \cite{zhao2010design} established a nonlinear observer to estimate vehicle velocity using a 3-degree-of-freedom vehicle model with a Dugoff tire model. Wenzel et al. \cite{wenzel2006dual} established dual extended Kalman filter (DEKF) parallel computations for simultaneous vehicle velocity estimation and parameter estimation, combining a 4-degree-of-freedom vehicle model with various tire models, and compared the estimation results of different combinations. Guo et al. \cite{guo2012implementation} proposed a field programmable gate array (FPGA) based on a system on programmable chip (SoPC) to improve the performance of EKF-based vehicle velocity estimation. Antonov et al. \cite{antonov2011unscented} developed an unscented Kalman filter (UKF) vehicle velocity observer based on a two-track model with the empiric magic formula. In addition, Rafatnia et al. \cite{rafatnia2021adaptive} corrected the vehicle velocity estimation based on a dynamics-based approach using low-frequency vehicle velocity measured by the global navigation satellite system (GNSS). However, the above dynamics-based estimation methods assume that the road conditions are already known. Therefore, other researchers have chosen to simultaneously estimate both the road friction coefficient and vehicle velocity. Magallan et al. \cite{magallan2010maximization} constructed a sliding-mode observer to estimate vehicle velocity based on the LuGre tire model using only the wheel speeds while estimating the road friction coefficient using other measurements. Zhang et al. \cite{zhang2014vehicle} used the braking torque, wheel speeds, and acceleration measurements as inputs to a sliding-mode observer to estimate vehicle velocity, and their observer used the Burckhardt tire model. The EKF was simultaneously established to estimate the friction parameters of the tire model. Han et al. \cite{han2017adaptive} concurrently estimated the road friction coefficient and vehicle velocity after the front wheel braking torques were saturated. The problems associated with these methods mainly lie in the requirement of accurate tire model parameters and the need to consider tire wear, inflation pressure, etc. To overcome these current problems, researchers have attempted a number of new approaches, taking road uncertainties and model parameter variations into account while developing observers. Meanwhile, kinematic-based and dynamics-based approaches have been combined to estimate the vehicle velocity \cite{hashemi2017corner,jalali2017integrated}. Based on this method, Pirani et al. \cite{pirani2017resilient} incorporated the effect of suspension compliance into the estimator to obtain a more accurate velocity at each corner (tire) for vehicles. Park et al. \cite{park2018integrated} developed an integrated EKF-based observer, one based on kinematics and one based on a bicycle model, with a weighting factor to select the appropriate method for the final velocity estimation in real time.

In addition to the above method of estimating the slip ratio by vehicle velocity, researchers have also used other parameters to replace the vehicle velocity, such as vehicle longitudinal acceleration or electric vehicle driving torque, together with the wheel angular velocity for calculations. Hori et al. \cite{hori1998traction} used the wheel velocity of the nondriven wheel instead of the vehicle velocity to estimate the slip ratio because the vehicle velocity is not easily available. However, this method fails in four-wheel drive vehicles and when braking. Later, Maeda et al. \cite{maeda2012four} used onboard acceleration sensors with wheel velocity to establish a slip ratio observer but encountered slow convergence and relatively large estimation errors. To improve the convergence of the estimator, Vo-Duy et al. \cite{vo2018slip} used the measured and estimated acceleration as the feedback term of the observer, which slightly improves convergence, but the estimated acceleration used in the feedback term has a simplification problem, while the noise and offset caused by the acceleration sensor still have an impact on the slip ratio estimation error. In addition to the abovementioned methods, there are other methods based on motor torque. Cecotti et al. \cite{cecotti2012estimation} added a small cosine signal to driving torque, which resulted in small variations in the wheel angular velocity and measured these variations to obtain the phase shift and gain relative to the torque oscillation; thus, the slip ratio was estimated based on the transfer function between the wheel angular velocity and driving torque. However, the results of this method are more sensitive to the frequency of the added signal. Boisvert et al. \cite{boisvert2016estimators} built an empirical nonlinear second-order system model of both slip ratio and motor torque; the model parameters need to be identified for specific operating conditions, but for unanticipated operating conditions, they need to be reidentified.

While the abovementioned methods can satisfy the existing body control stability systems, for future-oriented autonomous vehicles, we need to develop a new slip ratio estimation method with high accuracy, low latency, simplicity, and broader applicability. During the last decade, several sensing methods have been used to develop intelligent tires, such as microelectromechanical system (MEMS) acceleration sensors \cite{xu2020tire,xu2021tire}, stress sensors \cite{maurya20203d,maurya2018energy}, triboelectric nanogenerators (TENGs) \cite{askari2019tire,askari2018towards}, capacitive sensors \cite{matsuzaki2007wireless,matsuzaki2006passive}, optical fiber sensors \cite{coppo2017multisensing,roveri2016optyre}, surface acoustic wave sensors \cite{pohl1999intelligent,reindl2001saw}, magnetic sensors \cite{yilmazoglu2001integrated}, optical sensors \cite{tuononen2008optical} and ultrasonic distance sensors \cite{magori1998line}. Additionally, using intelligent tires, researchers have estimated tire forces \cite{xu2020tire}, slip angles \cite{xu2021tire} and road friction coefficients \cite{matsuzaki2014intelligent,niskanen2015three} with high accuracy. In this research, an intelligent tire system is developed based on an accelerometer to estimate the tire slip ratio. The three accelerations ($a_x$, $a_y$, and $a_z$) generated by the intelligent tire system are collected by a National Instrument (NI) data acquisition system (DAQ). The tire slip ratio reference signals are measured by the MTS tire testing system. Then, we investigate in detail the correlation between accelerations and slip ratio in three directions and find very useful features, especially vertical acceleration. These features are then trained by machine learning (ML) algorithms to build slip ratio estimation models. Different combinations of the three directional accelerations based on the vertical acceleration, which are ($a_x,a_y,a_z$), ($a_x,a_z$), ($a_y,a_z$), and $a_z$, are used as inputs to the algorithm. Additionally, to perform a comprehensive analysis and reliable estimation based on the ML-based slip ratio estimation algorithm using intelligent tires, four machine learning algorithms are tested: artificial neural networks (ANNs), gradient boosting machines (GBMs), random forests (RFs), and support vector machines (SVMs).

The framework of this paper is organized as follows. Section \ref{sec:02} introduces the test system as well as the design of the working conditions, where the test system includes the Measure Test Simulate (MTS) tire test system and the intelligent tire system. Section \ref{sec:03} analyzes the correlation between the accelerations and the slip ratio. Section \ref{sec:04} presents the data preprocessing for training machine learning techniques. Section \ref{sec:05} comprehensively evaluates the estimation results of the ML algorithms using four inputs for the slip ratio. In addition, a 10-fold cross-validation (CV) is provided to study the accuracy of four ML-based slip ratio estimation algorithms. Finally, the conclusions and future research work of this paper are stated.
\section{Testing Platform and Data Acquisition System}
\label{sec:02}
\begin{figure}[!h]
    \centering
    \includegraphics[width=3in]{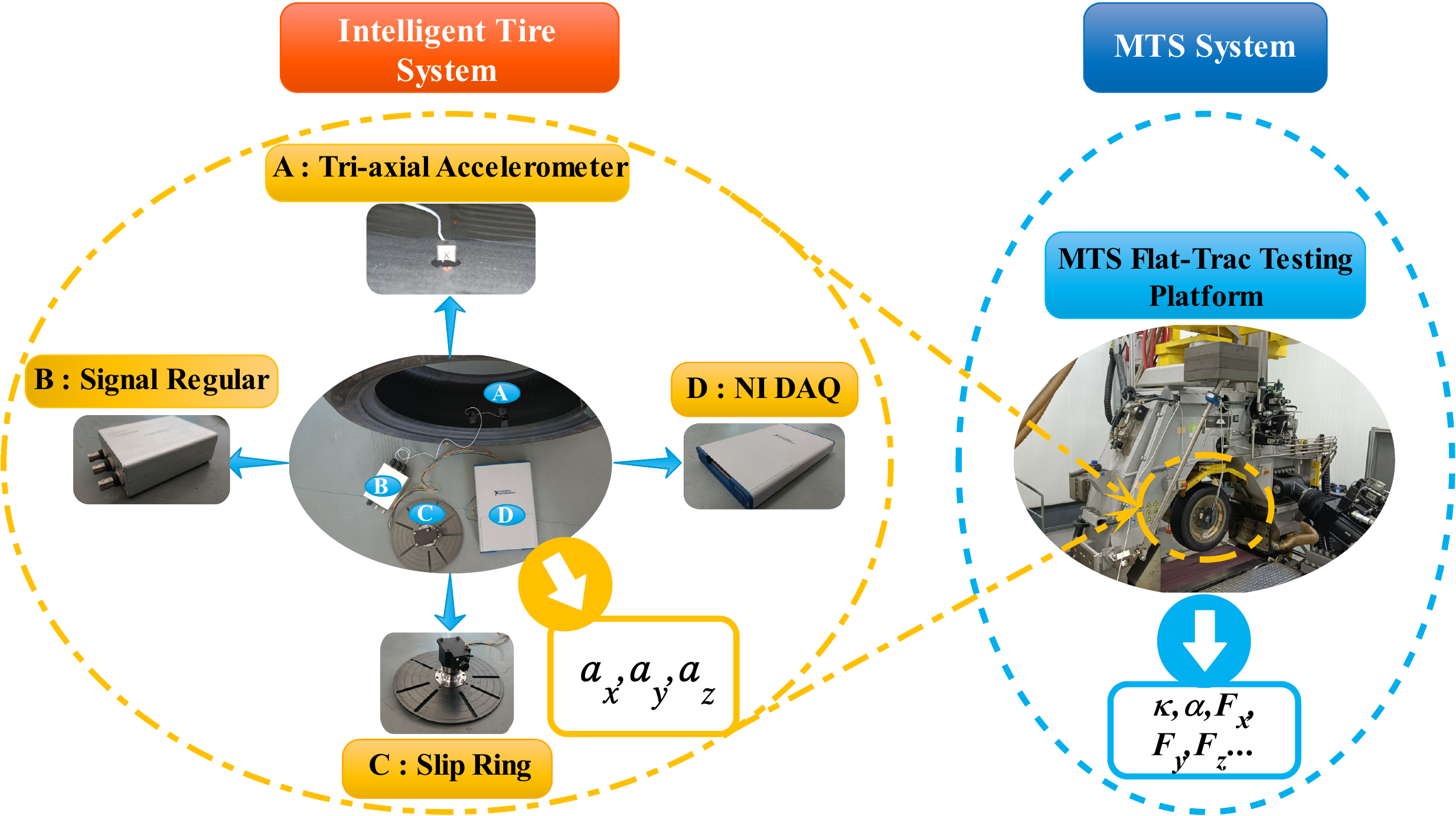}
\caption{The entire test system.}
    \label{fig:entire test}
\end{figure}

The testing platform includes two main parts, namely, the Measure Test Simulate (MTS) tire testing system and the intelligent tire module. The MTS tire test system is used to set up different driving scenarios for a wide range of slip ratios, slip angles and tire forces. The tire contact patch accelerations are extracted with the use of an accelerometer attached to the intelligent tire. The whole testing system is shown in Fig. \ref{fig:entire test}.

\subsection{MTS Tire Experimental System}

\begin{figure}[!h]
    \centering
    \includegraphics[width=3in]{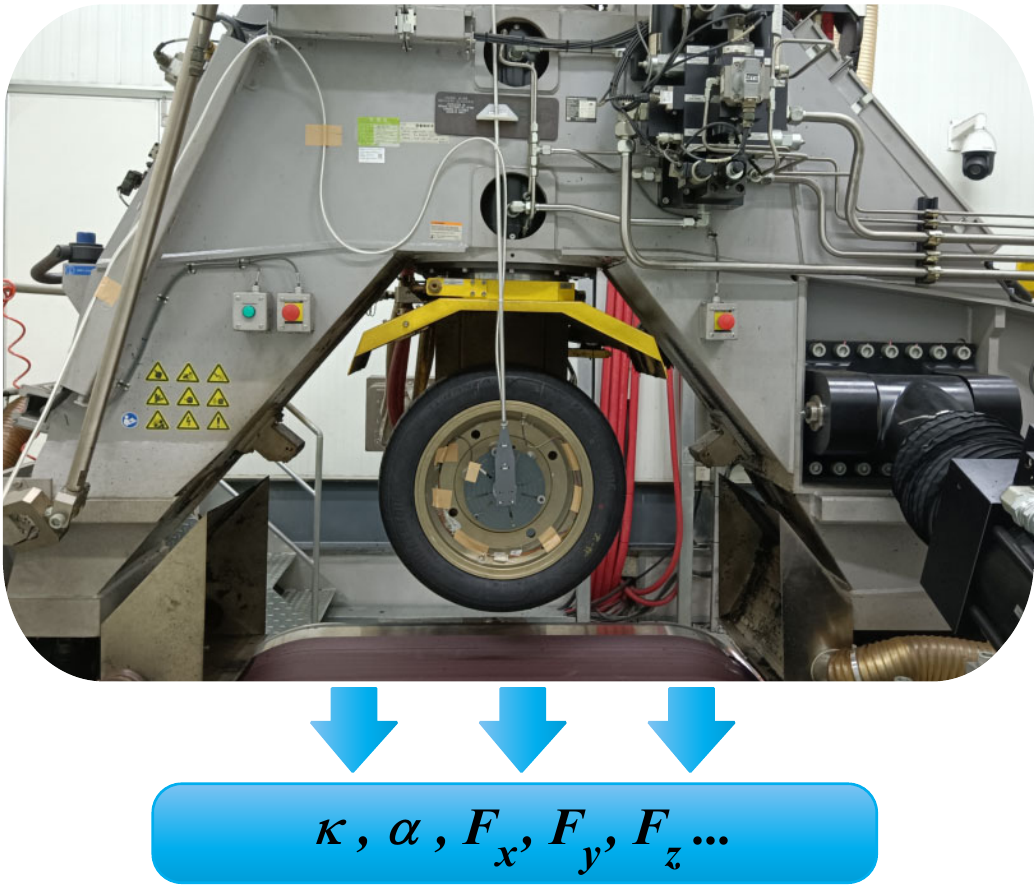}
\caption{MTS tire testing platform.}
    \label{fig:MTS}
\end{figure}

To extract the slip ratio data, the MTS testing apparatus is used to experimentally simulate different maneuvers, see Fig. \ref{fig:MTS}. With the use of this advanced testing module, we can obtain the tire slip ratio ($\kappa$), slip angle ($\alpha$), longitudinal force ($F_x$), lateral force ($F_y$), and vertical force ($F_z$) on a real-time basis, as shown in Fig. \ref{fig:MTS}. These data can be used for validating the estimation results based on intelligent tire data.

\subsection{Intelligent Tire System}
The intelligent tire system entails a triaxial acceleration sensor, a high-speed slip ring, a signal regulator, and a National Instrument (NI) data acquisition system (DAQ), as shown in Fig. \ref{fig:entire test}.

\begin{figure}[htb]
    \centering
\begin{minipage}[t]{0.45\linewidth}
    \centering
    \includegraphics[scale=0.5]{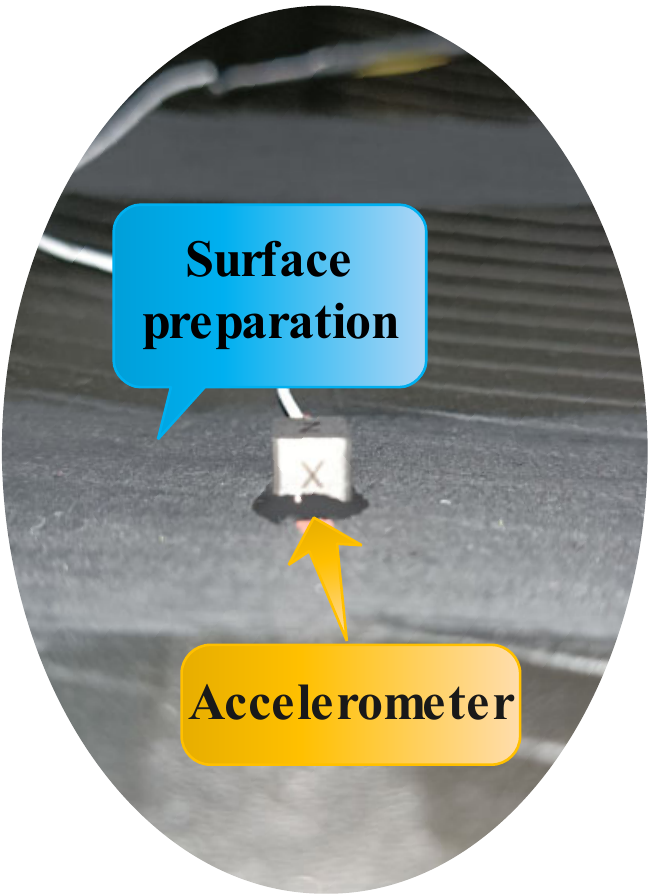}
\subcaption{}
    \label{fig:set up acc.a}
\end{minipage}
\begin{minipage}[t]{0.45\linewidth}
    \centering
    \includegraphics[scale=0.5]{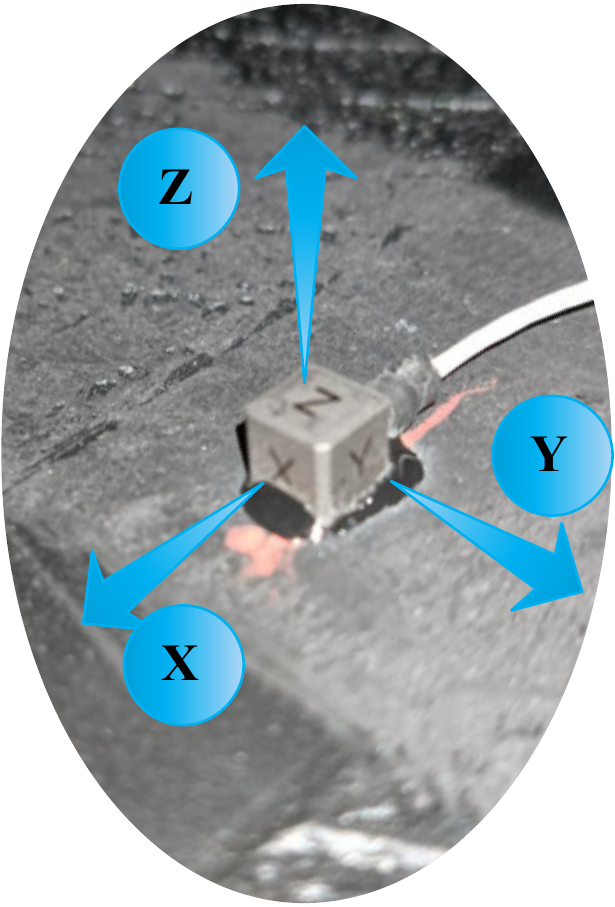}
\subcaption{}
    \label{fig:set up acc.b}
\end{minipage}
\caption{(a) Installation of accelerometer in tire; (b) accelerometer coordination system within tire.}
    \label{fig:set up acc}
\end{figure}

\begin{figure}[!h]
    \centering
\begin{minipage}[t]{0.45\linewidth}
    \centering
    \includegraphics[scale=0.3]{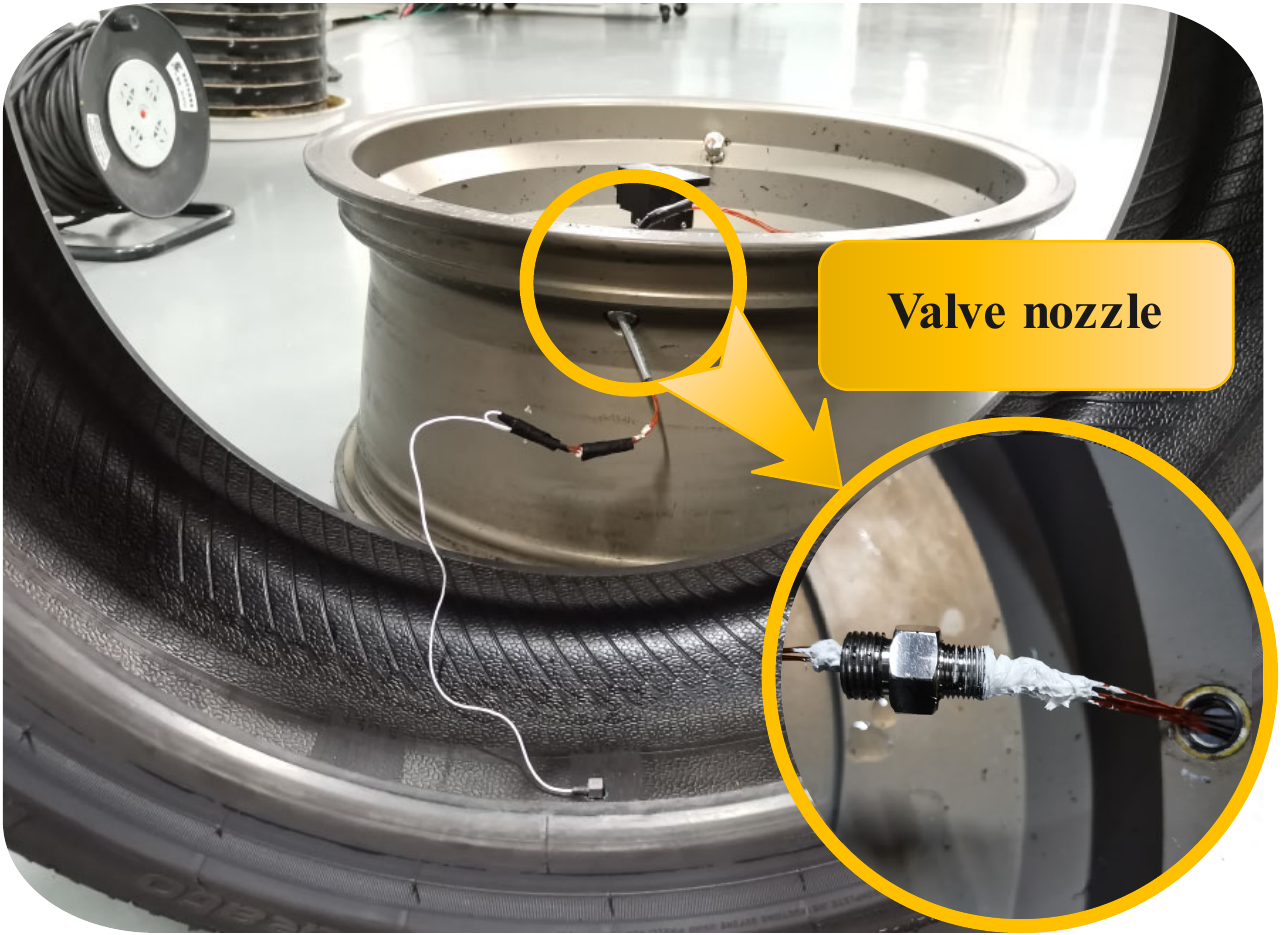}
\subcaption{}
    \label{fig:acc sensor.a}
\end{minipage}
\begin{minipage}[t]{0.45\linewidth}
    \centering
    \includegraphics[scale=0.3]{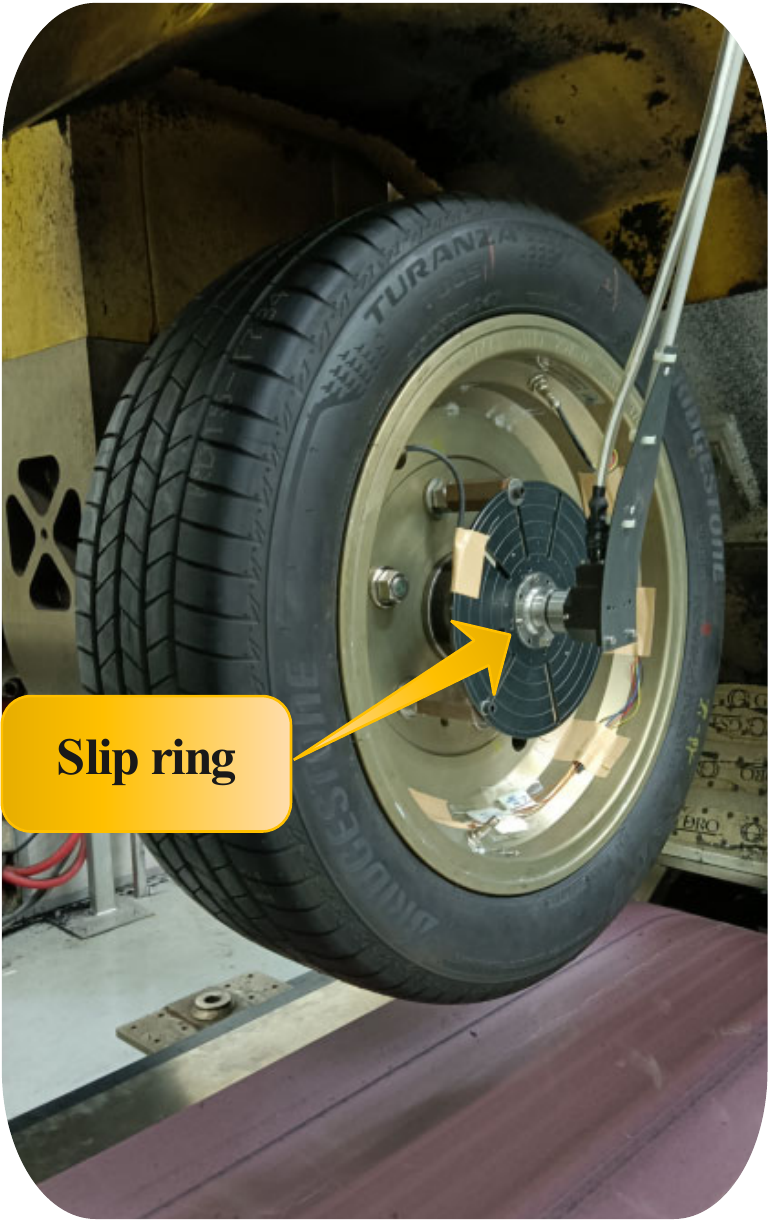}
\subcaption{}
    \label{fig:acc sensor.b}
\end{minipage}
\caption{Sensor wiring passes through (a) the valve nozzle and (b) slip ring.}
    \label{fig:acc sensor}
\end{figure}

The triaxial acceleration sensor is glued to the inner liner of the tire to measure the accelerations ($a_x$, $a_y$, and $a_z$) generated in the tire contact patch, as shown in Fig. \ref{fig:set up acc}(a). The figure shows that the sensor mounting surface is pretreated to have more reliable adhesion. The considered coordination system is shown in Fig. \ref{fig:set up acc}(b), where the $x$-direction is consistent with the forward direction of the tire. The sensor wiring passes through the valve nozzle of the rim (see Fig. \ref{fig:acc sensor}(a)), where the valve nozzle device is sealed with sealant to avoid tire air leakage. After exiting the rim, it has to pass through the high-speed slip ring in Fig. \ref{fig:acc sensor}(b), which is designed to be mounted on the rim and used to transmit the accelerations from the rotating tire. The signal regulator uses three channels (corresponding to the three acceleration directions) to provide a constant current source to provide electrical power for the accelerometer. The NI DAQ system is used to collect the accelerations by adjusting the signal channels, sampling frequency and sampling method. The single-ended reference mode (selected when the signal regulator is grounded) is adopted for this test. Additionally, a sampling frequency of 10 kHz is chosen, which meets the research requirements.

\subsection{Test Scenarios}

We use tires manufactured by Bridgestone for the experiments carried out in this study. We consider three different forms of loads, two speeds, and different forms of slip ratios in the experimental cases, as shown in Table \ref{tab:my-table-01}. In our experiments, we consider two forms of variations for the tire slip ratio, covering the drive/braking conditions in daily driving: one with small values (Data Set 1 and 2), which occur mainly in our normal driving conditions, and the tire is in the linear region; the other with large values (Data Set 3), which occur mainly when the vehicle is handling emergency conditions. At this time, the tire is in the nonlinear region and suffers severe deformation affecting the acceleration. The complex mechanical properties also pose a challenge for the estimation of the large-value slip ratio. The data are used to fully evaluate the slip ratio estimation model.

\begin{table}[!h]
\centering
\caption{\MakeUppercase{Testing Scenarios and sample classification}}
\label{tab:my-table-01}
\begin{tabular}{ccl}
\cline{1-2}
\multicolumn{2}{c}{\textbf{Testing scenarios}}                                                                                          &  \\ \cline{1-2}
\textbf{Driving/braking} & \textbf{Parameters}                                                                                          &  \\
Tire brand               & Bridgestone 215/55R17                                                                                        &  \\
Pressure {[}kPa{]}       & 250                                                                                                          &  \\
Vertical load {[}N{]}    & 2680 4020 5360                                                                                               &  \\
Velocity {[}km/h{]}      & 30 60                                                                                                        &  \\
Slip ratio {[}\%{]}      & \begin{tabular}[c]{@{}c@{}}$\pm4,\pm3,\pm2,\pm1$, triangular wave (amplitude 3 and\\  up to 30)\end{tabular} &  \\ \cline{1-2}
\multicolumn{2}{c}{\textbf{Sample classification}}                                                                                      &  \\ \cline{1-2}
\textbf{Datasets}        & \textbf{Parameters}                                                                                          &  \\
Data Set 1               & $\pm4,\pm3,\pm2,\pm1$                                                                                        &  \\
Data Set 2               & Triangular wave (amplitude 3)                                                                                &  \\
Data Set 3               & Triangular wave (amplitude 30)                                                                               &  \\
Data Set 4               & Collection of Data Set 1,2,3        
                                                                 &  \\
\cline{1-2}
\end{tabular}
\end{table}

In Data Set 1, we consider that the slip ratio remains constant during certain tire rotations, and then it gradually increases from -4\% to 4\%. Data Sets 2 and 3 are based on a continuous variation in the slip ratio considering a triangular form. Data Set 2 includes a small portion of the continuous variation in the slip ratio with a 3\% amplitude. Data Set 3 considers a larger range of slip ratios with 30\% amplitude. Considering all three data sets, the experimental data include 2218 tire revolutions (1082 in Data Set 1, 666 in Data Set 2, and 470 in Data Set 3). All of these data are used to train four different machine learning algorithms to estimate the tire slip ratio. Each of them is divided into training (70\%) and testing datasets (30\%).
\section{Data Analysis}
\label{sec:03}
The focus of this section is on finding the correlation between tire contact patch accelerations and the slip ratio. The accelerations in the figures below are filtered by 400 Hz Butterworth low-pass filtering \cite{xu2020tire}.

\begin{figure}[!h]
    \centering
    \includegraphics[width=3in]{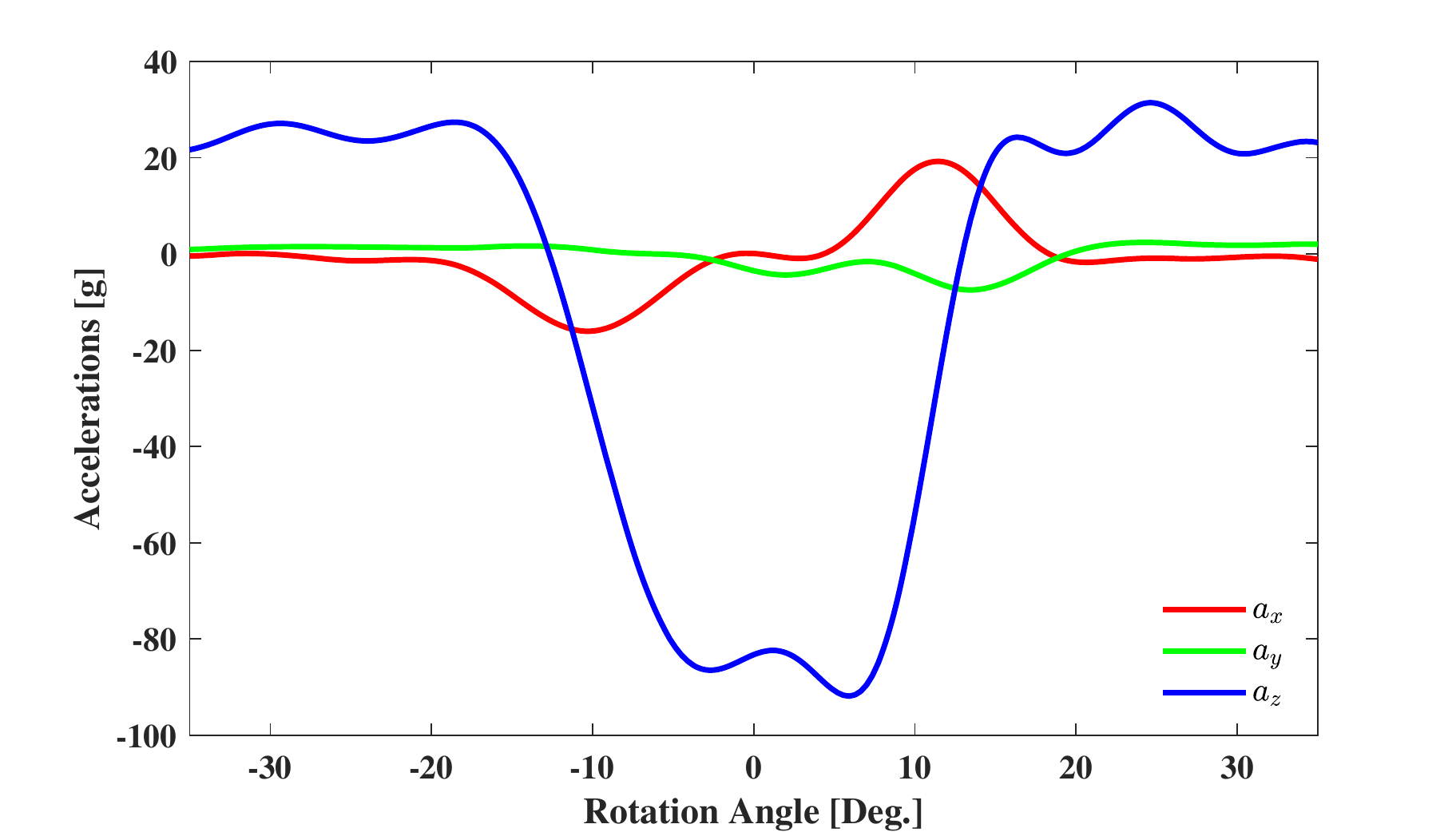}
\caption{Acceleration with a 0\% slip ratio (60 km/h at 4020 N vertical load).}
    \label{fig:three direction}
\end{figure}

First, the basic accelerations in three directions are given, as shown in Fig. \ref{fig:three direction}, which shows the free rolling condition at 60 km/h under 4020 N. In particular, the longitudinal acceleration and the vertical acceleration show symmetric distributions under this condition. In the figure, the longitudinal acceleration includes two peaks while entering and leaving the contact patch, the vertical acceleration is a clear feature of the low center and high sides, and the lateral acceleration has no significant change because there is no lateral motion in the free rolling condition. These characteristics of acceleration are consistent with previous studies \cite{xu2020tire,jeong2021tire}.

\begin{figure}[!h]
    \centering
\begin{minipage}[t]{0.45\linewidth}
    \centering
    \includegraphics[scale=0.52]{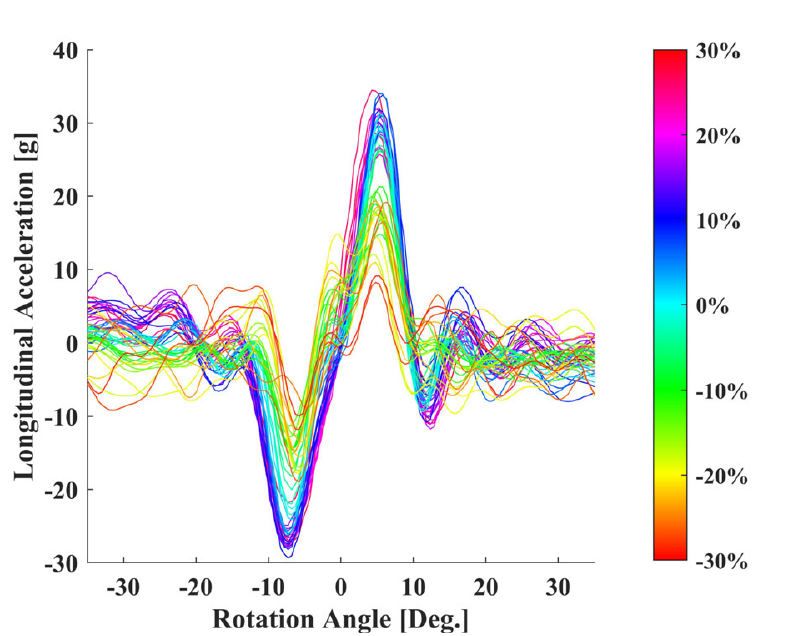}
\subcaption{}
\end{minipage}
\begin{minipage}[t]{0.45\linewidth}
    \centering
    \includegraphics[scale=0.52]{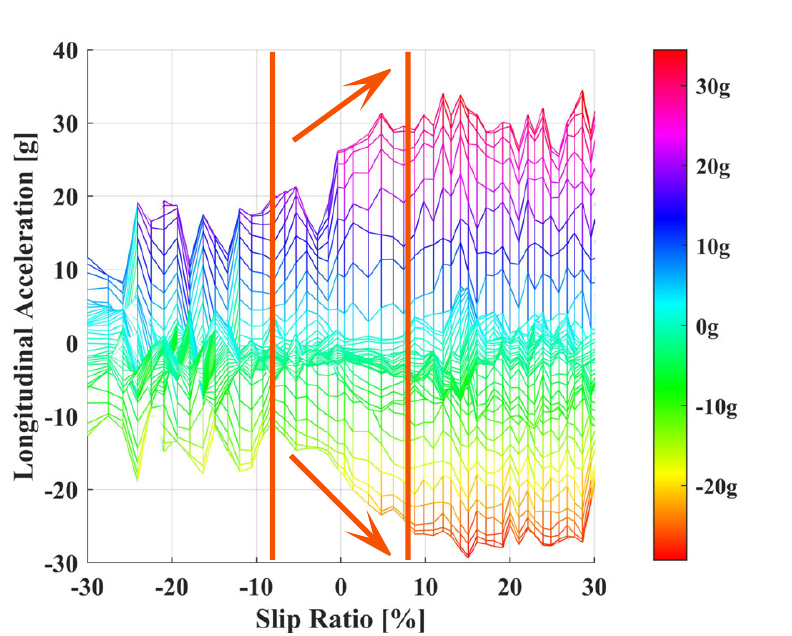}
\subcaption{}
\end{minipage}
\caption{Longitudinal acceleration (Data Set 3 - 60 km/h at 2680 N vertical load).}
    \label{fig:csr accx}
\end{figure}

\begin{figure}[!h]
    \centering
\begin{minipage}[t]{0.45\linewidth}
    \centering
    \includegraphics[scale=0.52]{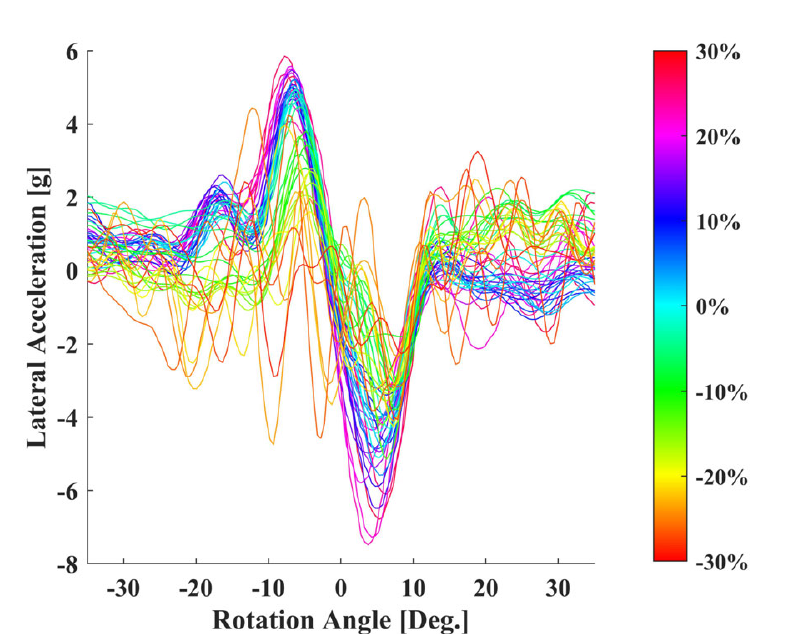}
\subcaption{}
\end{minipage}
\begin{minipage}[t]{0.45\linewidth}
    \centering
    \includegraphics[scale=0.52]{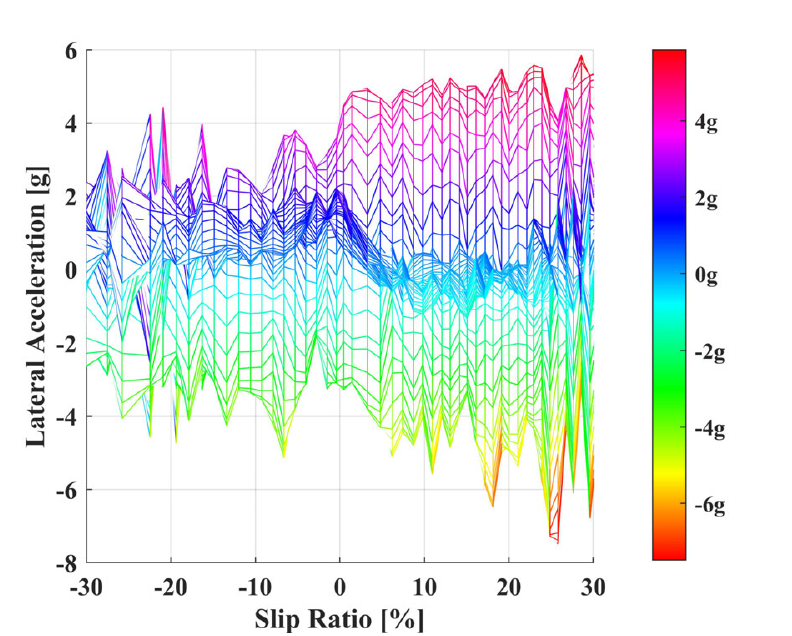}
\subcaption{}
\end{minipage}
\caption{Lateral acceleration (Data Set 3 - 60 km/h at 2680 N vertical load).}
    \label{fig:csr accy}
\end{figure}

\begin{figure}[!h]
    \centering
\begin{minipage}[t]{0.45\linewidth}
    \centering
    \includegraphics[scale=0.52]{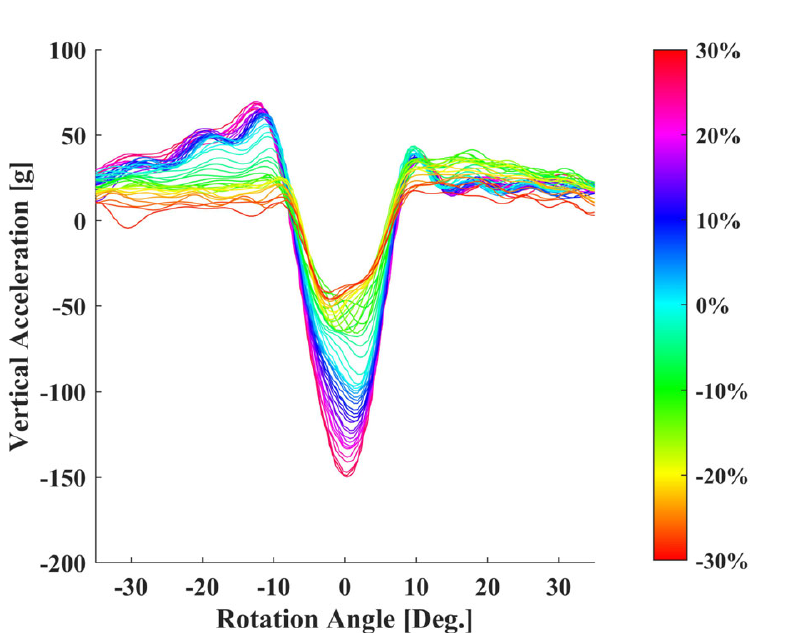}
\subcaption{}
\end{minipage}
\begin{minipage}[t]{0.45\linewidth}
    \centering
    \includegraphics[scale=0.52]{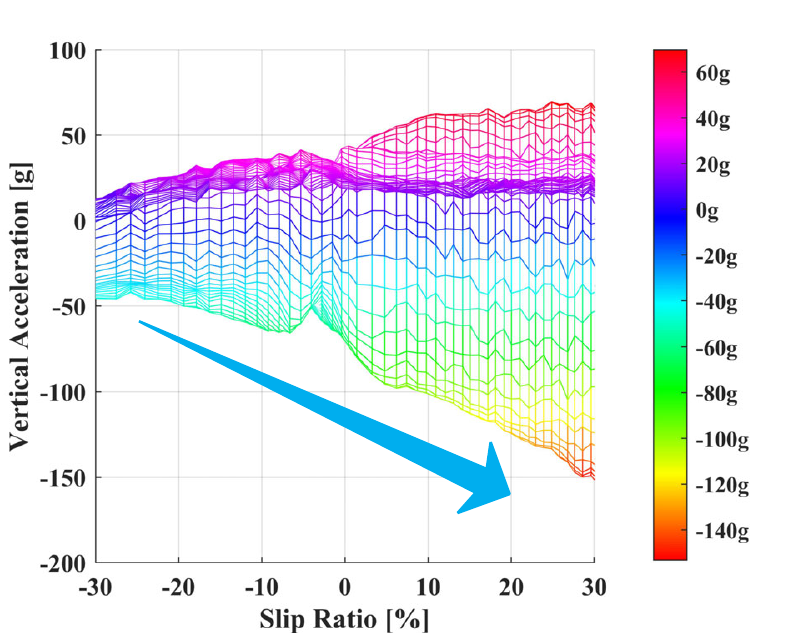}
\subcaption{}
\end{minipage}
\caption{Vertical acceleration (Data Set 3 - 60 km/h at 2680 N vertical load).}
    \label{fig:csr accz}
\end{figure}

\begin{figure}[!h]
    \centering
    \includegraphics[width=3in]{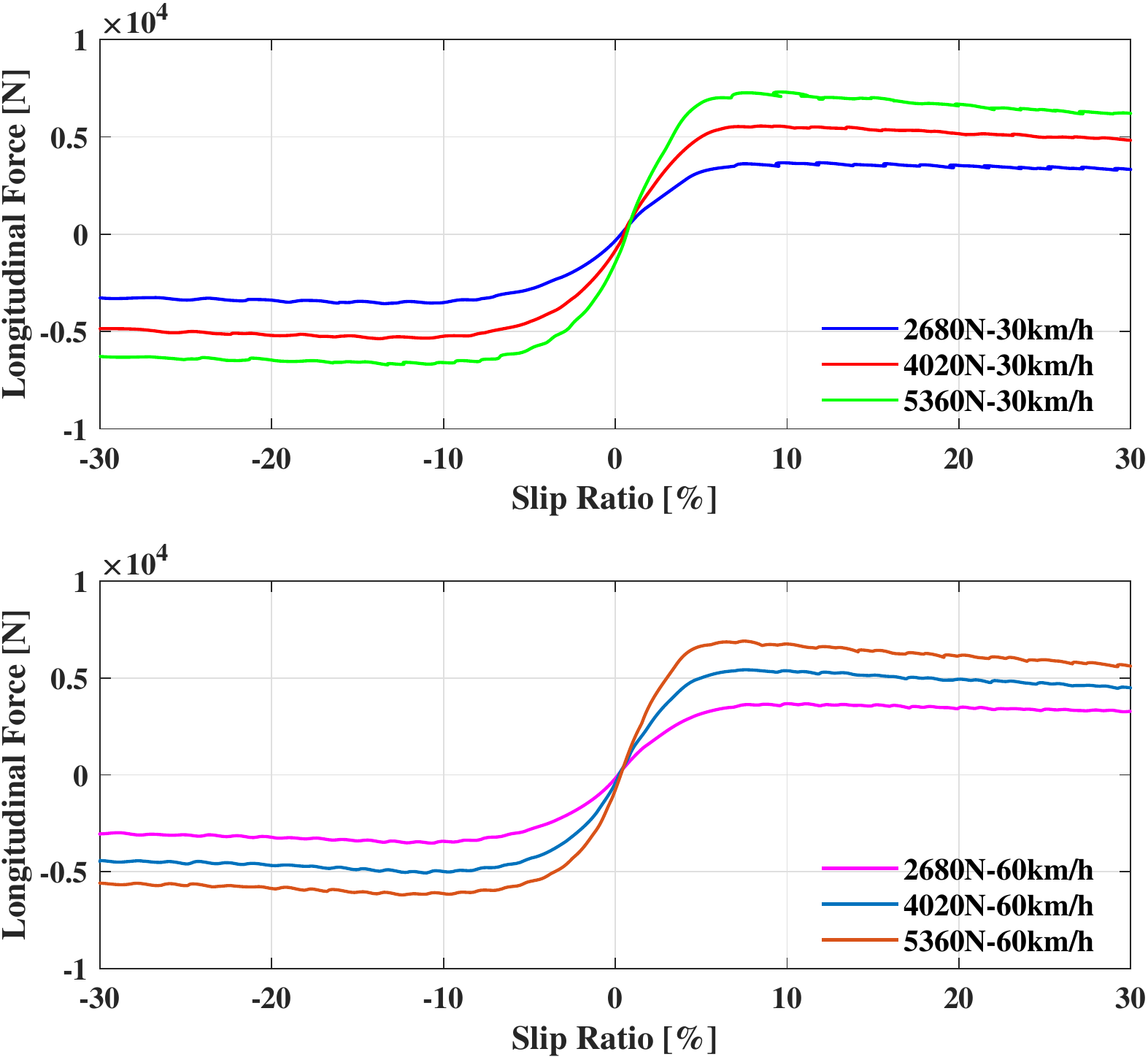}
\caption{Longitudinal force vs. slip ratio under different working conditions.}
    \label{fig:Fx_SR}
\end{figure}

Next, we explore the features of the three directions of acceleration that occur when the tire slips. Figs. \ref{fig:csr accx}-\ref{fig:csr accz} show $a_x$, $a_y$ and $a_z$ using Data Set 3, respectively. Each plot consists of two subplots: the left plot shows the tire rotation angle vs. acceleration window using a color bar of slip ratio, and the right plot shows the slip ratio vs. acceleration window using a color bar of acceleration. In Fig. \ref{fig:csr accx}(a), we can observe two peaks (same as in Fig. \ref{fig:three direction}). Additionally, their magnitudes, which can be observed in combination with the right subplot, vary with the slip ratio. A more detailed description is presented below: the absolute values of maximum and minimum longitudinal acceleration increase linearly as the slip ratio growths from -8\% to 8\% in Fig. \ref{fig:csr accx}(b). This feature is dependent on the wheel velocity. When the absolute value of the slip ratio exceeds 8\%, they become saturated. This feature is similar to the nonlinear variation relationship between the longitudinal slip ratio vs. the longitudinal force shown in Fig. \ref{fig:Fx_SR}.

Based on Fig. \ref{fig:csr accy}(a), it is observed that the lateral acceleration behaves in opposite order compared to longitudinal acceleration. When entering and leaving the tire contact patch, lateral acceleration increases first and then decreases. Additionally, the data analyzed are for longitudinal slip conditions, which have no motion in the lateral direction of the tire. Therefore, there does not seem to be a certain correlation between lateral acceleration and slip ratio based on Fig. \ref{fig:csr accy}(b).

\begin{figure}[!h]
    \centering
    \includegraphics[width=3in]{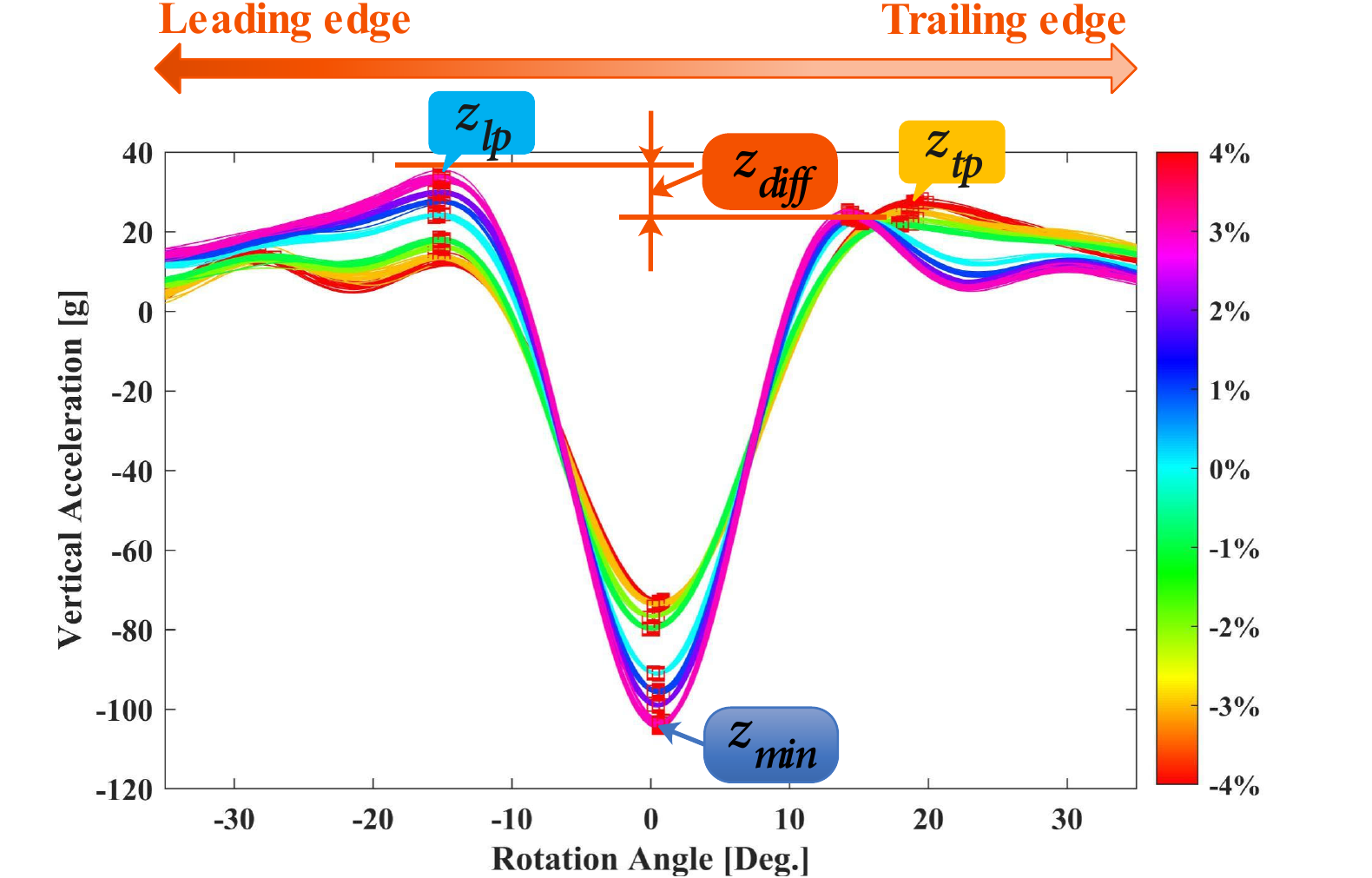}
\caption{Vertical acceleration (Data Set 1 - 60 km/h at 2680 N vertical load).}
    \label{fig:ssr accz}
\end{figure}

In comparison to the longitudinal and vertical acceleration, the vertical acceleration can be seen in Fig. \ref{fig:csr accz} with distinct features, such as the leading-trailing peak variation in Fig. \ref{fig:csr accz}(a) and the minimum vertical acceleration in Fig. \ref{fig:csr accz}(b). The minimum vertical acceleration shows a linear variation with the slip ratio. For a clearer expression, the accelerations at a small slip ratio of Data Set 1 are chosen to label these features in Fig. \ref{fig:ssr accz}. The minimum vertical acceleration in the center is labeled $z_{min}$, and the leading-trailing peaks are labeled $z_{lp}$ and $z_{tp}$, respectively. In Fig. \ref{fig:ssr accz}, the positive rotation angle region represents the trailing edge of the tire contact patch, and the negative rotation angle region represents the leading edge of the tire contact patch. An attractive feature is that $z_{lp}$ is greater than $z_{tp}$ at a positive slip ratio and $z_{lp}$ is less than $z_{tp}$ at a negative slip ratio. This difference of $z_{lp}$ and $z_{tp}$ is characterized by $z_{diff}$. The features of $z_{min}$ and $z_{diff}$ in Fig. \ref{fig:ssr accz} at small slip ratios can also be seen in Fig. \ref{fig:csr accz} at large slip ratios.

\begin{figure}[!h]
    \centering
    \includegraphics[width=3in]{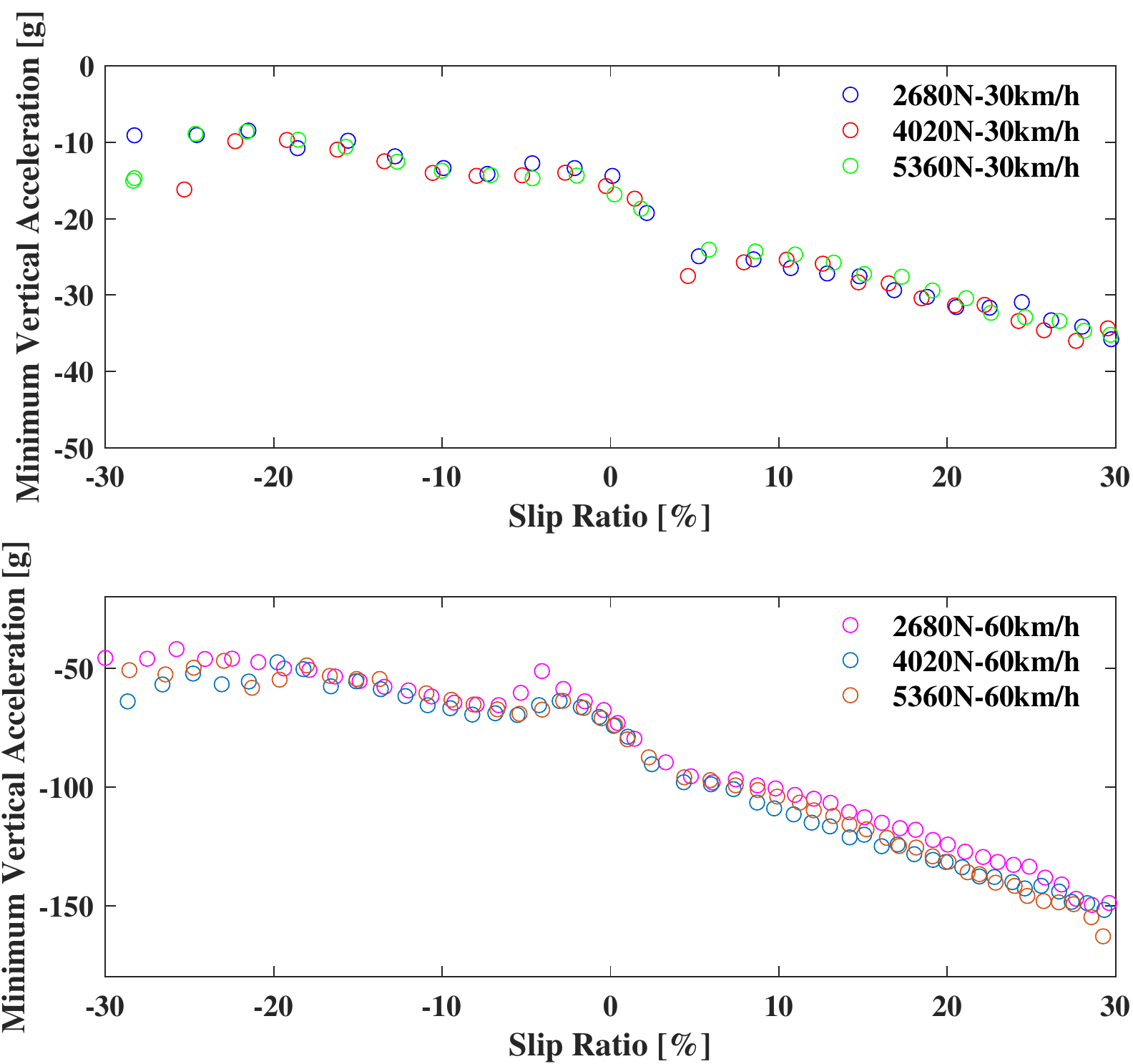}
\caption{The minimum vertical acceleration $z_{min}$ vs. slip ratio under different working conditions.}
    \label{fig:Min accz}
\end{figure}

\begin{figure}[!h]
    \centering
    \includegraphics[width=3in]{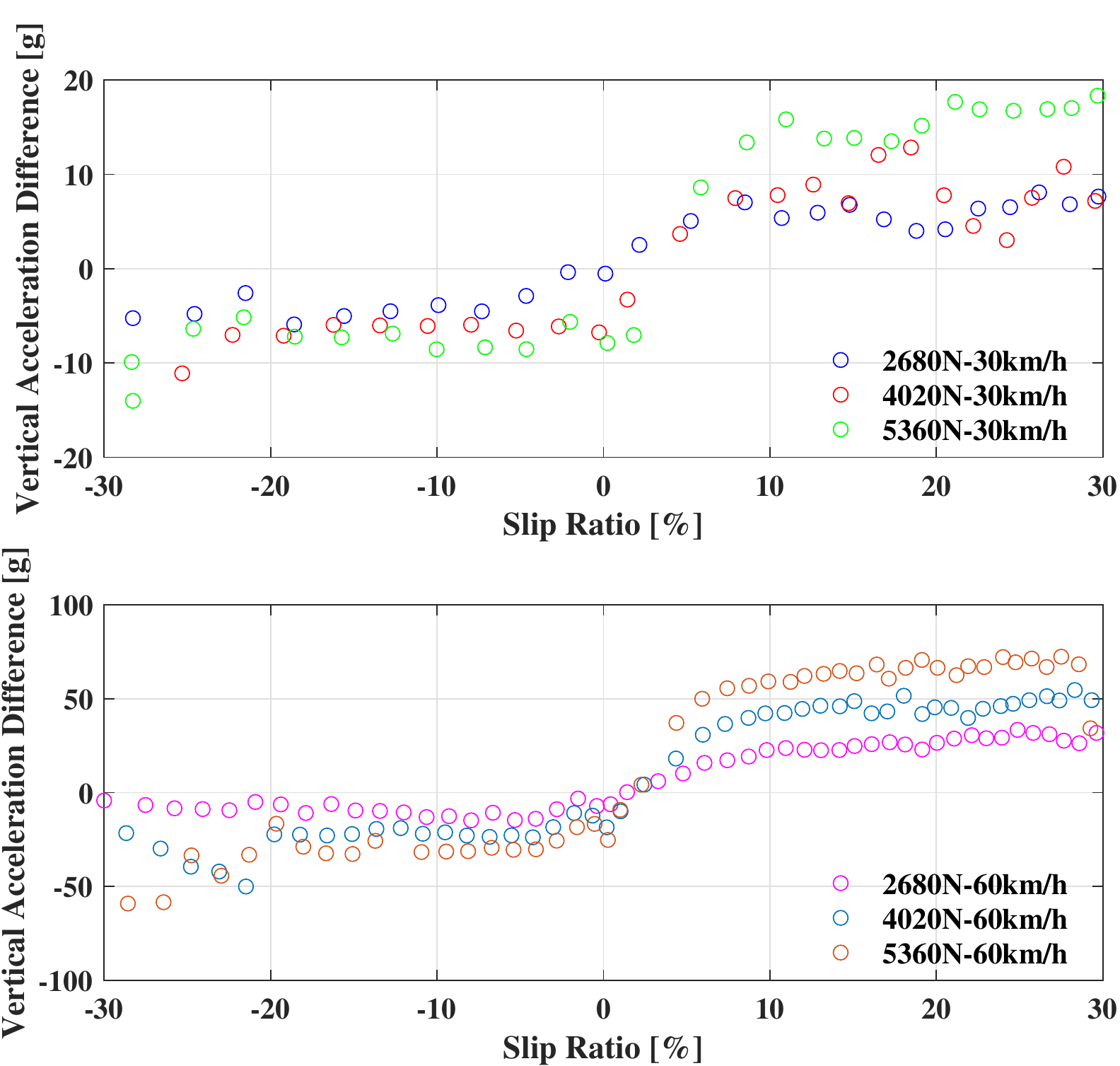}
\caption{The vertical acceleration difference $z_{diff}$ vs. slip ratio under different working conditions.}
    \label{fig:Dif accz}
\end{figure}

In summary, the longitudinal acceleration, although it has a relatively better feature with the slip ratio, does not behave reliably in other working conditions. The vertical acceleration shows a more stable and reliable feature and relation with the slip ratio. To further study the behavior of vertical acceleration, we plotted the values of $z_{min}$ and $z_{diff}$ in all experimental conditions, as shown in Figs. \ref{fig:Min accz} and Fig. \ref{fig:Dif accz}.

Fig. \ref{fig:Min accz} shows the minimum vertical acceleration vs. slip ratio. The figure is divided into two subplots, the top and bottom, which are related to 30 km/h and 60 km/h, respectively. It can be clearly seen that the variation in $z_{min}$ is linear with the slip ratio. As the tire speed decreases during the braking condition, the absolute value of $z_{min}$ is smaller compared to the driving condition. The absolute value of $z_{min}$ at 30 km/h is less than that at 60 km/h. Accordingly, $z_{min}$ is mainly affected by the wheel velocity. In addition, there is a linear relationship between $z_{min}$ and wheel velocity.

As Fig. \ref{fig:Dif accz} shows, $z_{diff}$ increases linearly between the -4\% and 8\% slip ratios. As the slip ratio exceeds the abovementioned range, $z_{diff}$ tends to saturate, and its growth rate slows down, resulting in a nonlinear relationship between the slip ratio and $z_{diff}$. In addition, Fig. \ref{fig:Fx_SR} shows the variation of slip ratio vs. longitudinal force. We also observe nonlinear behavior, which is based on tire mechanical properties \cite{guo2007unitire,pacejka1992magic}. In addition, we observe that the absolute value of $z_{diff}$ for a given slip ratio is higher for the positive range than for the negative range. This feature can be also identified in Fig. \ref{fig:Fx_SR}.

\begin{figure}[h!]
    \centering
\begin{minipage}[t]{0.45\linewidth}
    \centering
    \includegraphics[scale=0.5]{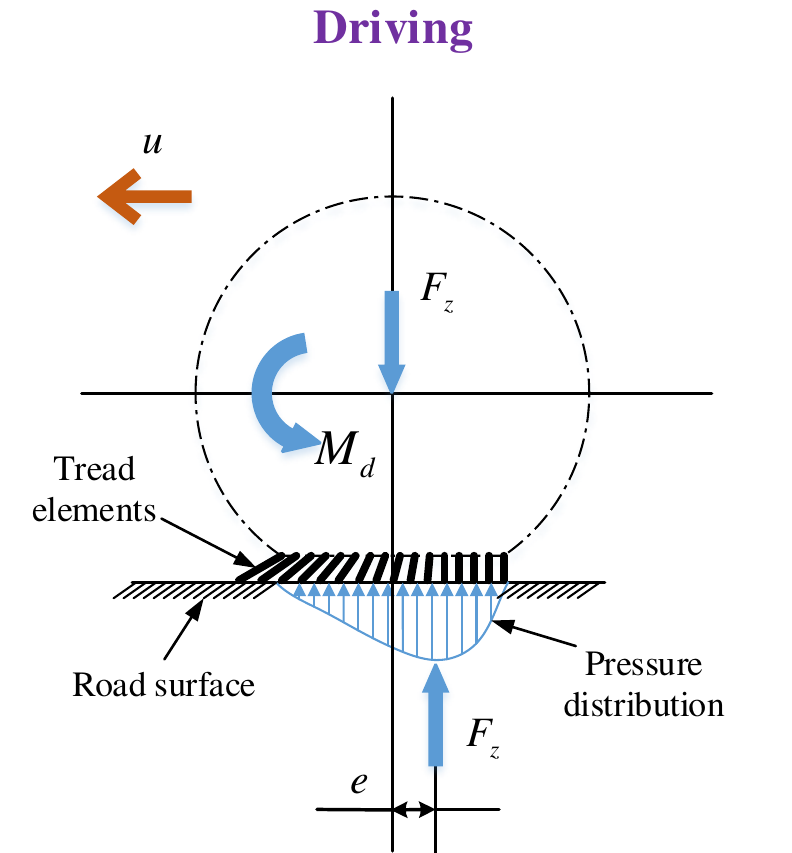}
\subcaption{}
    \label{fig:driving_braking.a}
\end{minipage}
\begin{minipage}[t]{0.45\linewidth}
    \centering
    \includegraphics[scale=0.5]{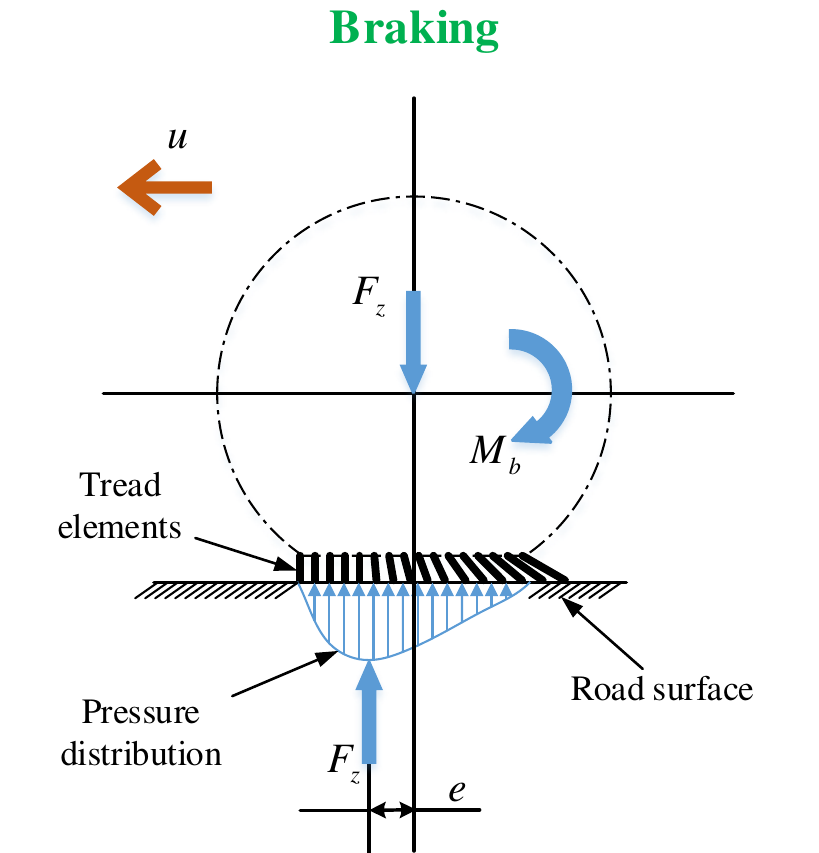}
\subcaption{}
    \label{fig:driving_braking.b}
\end{minipage}
\caption{Deformation of a tire element and pressure distribution in a tire contact patch: (a) driving condition; (b) braking condition.}
    \label{fig:driving_braking}
\end{figure}

\begin{figure}[!h]
    \centering
    \includegraphics[width=3in]{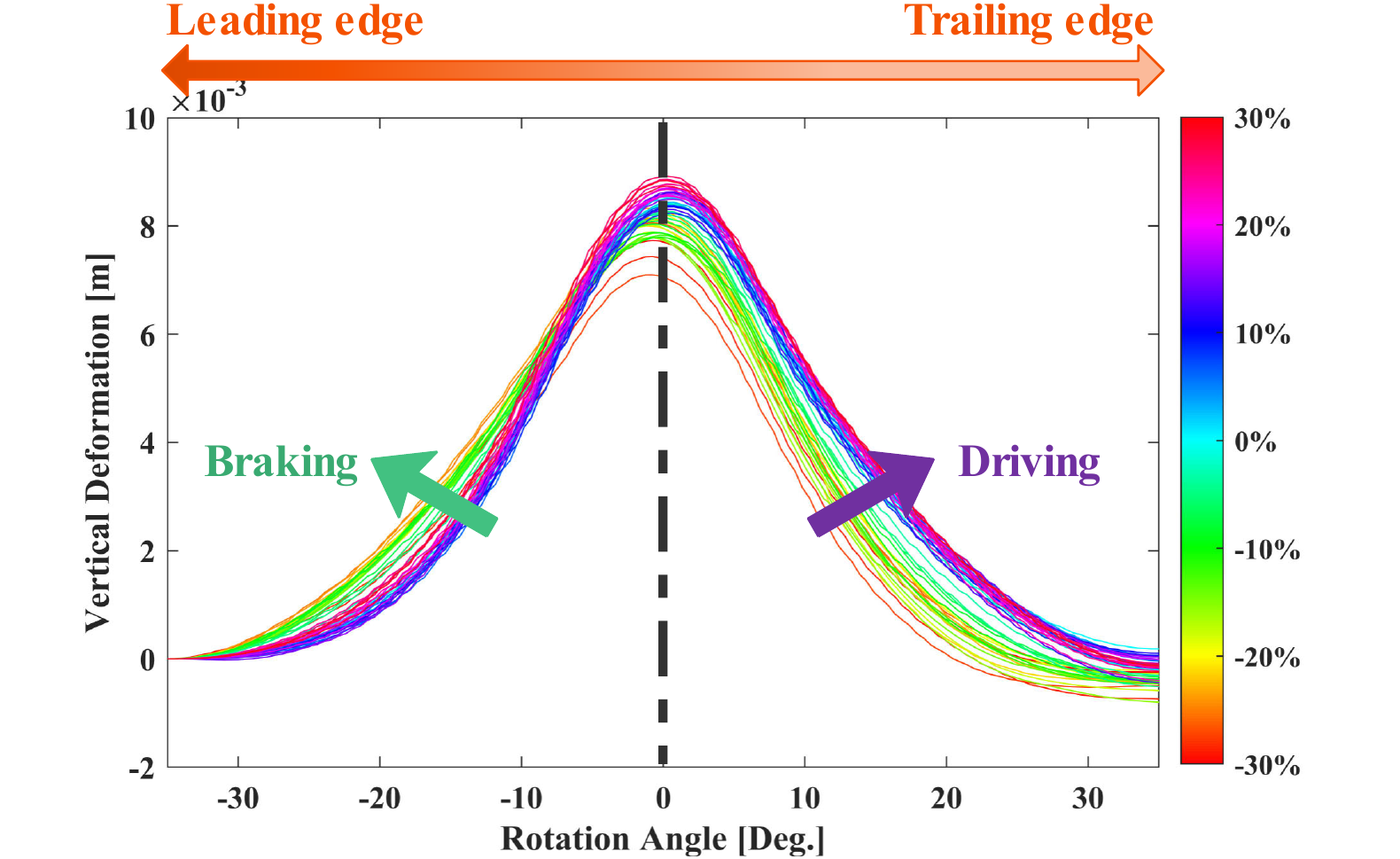}
\caption{Integral vertical deformation under driving and braking conditions.}
    \label{fig:Dz_SR}
\end{figure}

Based on the tire dynamics, the behavior of vertical acceleration with the slip ratio can be explained by the shape of the vertical deformation curves \cite{guo2007unitire}. Fig. \ref{fig:driving_braking} shows the tire tread element deformation and vertical pressure distribution under driving and braking conditions. $M_d$ and $M_b$ indicate the driving torque and braking torque, respectively. $F_z$ represents the vertical load, and $e$ is the shift of the gravity center of contact pressure. The longitudinal deformation of the tread elements within the contact patch is caused by slip, which gradually increases as the contact patch is continuously changed. In the case of driving, the vertical pressure distribution inside the contact patch is shifted towards the rear part of the contact patch center, and the maximum vertical deformation occurs at the trailing of the contact patch. In the case of braking, the maximum vertical deformation occurs at the leading of the contact patch as the increase in braking force is shifted towards the front part of the contact patch. This offset increases linearly with respect to the slip ratio until reaching a certain value. Accordingly, there will be neither an increase nor a decrease as the tread element deformation has reached its limit. We obtain the vertical deformation by double integration of the vertical acceleration, as shown in Fig. \ref{fig:Dz_SR}, which is characterized by the offset of the vertical deformation toward the trailing edge of the center of the contact patch when driving and toward the leading edge of the center of the contact patch when braking. This feature is the same as what has already been explained based on \ref{fig:driving_braking}.

In summary, the data analysis shows that there are promising features of longitudinal acceleration and vertical acceleration vs. slip ratio. We also analyzed the vertical acceleration using the tire dynamics. Next, we use machine learning algorithms to train the above features to the slip ratio estimation model.
\section{Methodology}
\label{sec:04}
In this section, it is shown how the extracted information from the tire contact patch is used to find the tire slip ratio. In addition, a detailed explanation is provided regarding the implemented ML techniques and adjusting their hyperparameters.

\subsection{Data Pre-Processing}
\begin{figure}[!h]
    \centering
    \includegraphics[width=3in]{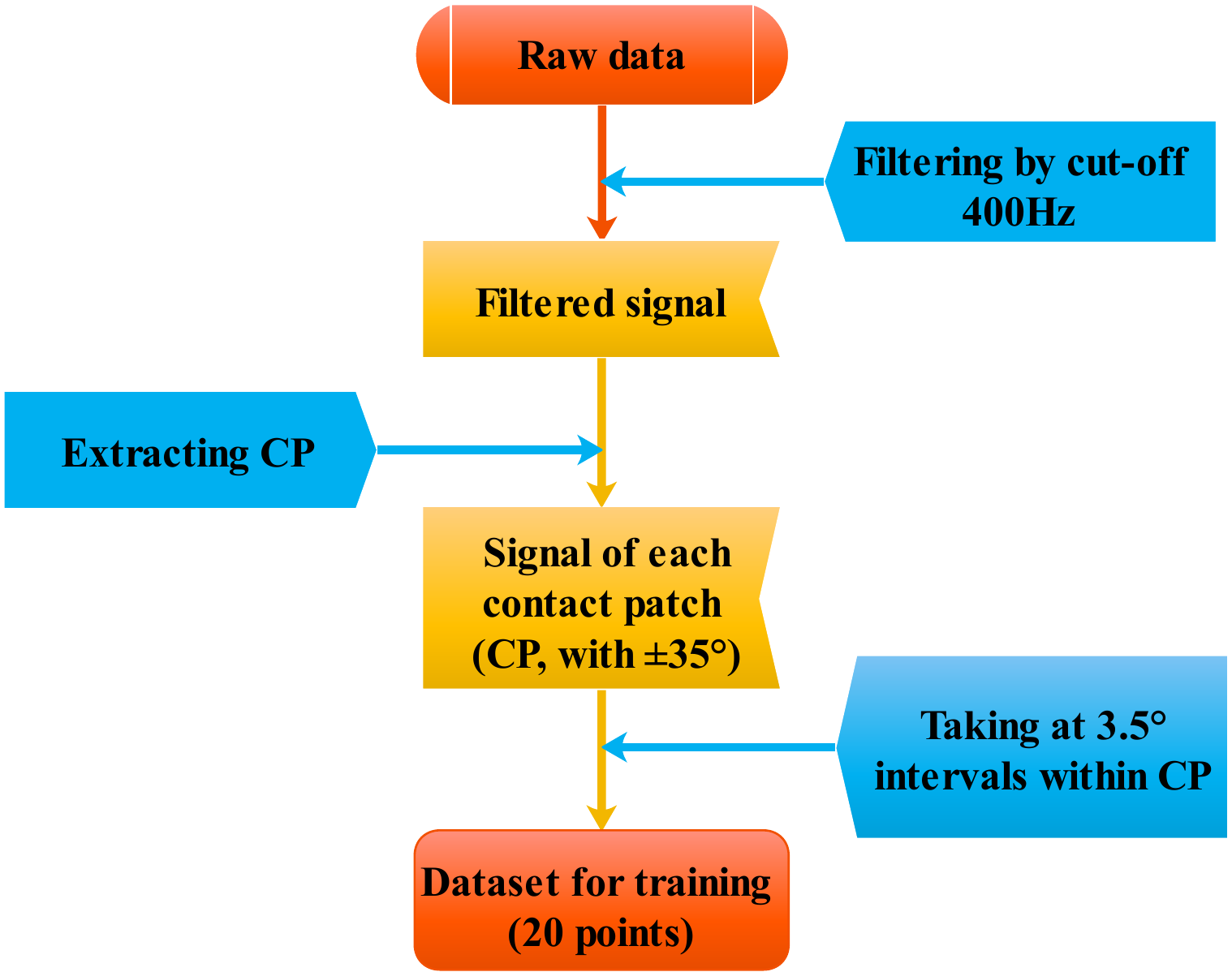}
\caption{Flowchart of data preprocessing.}
    \label{fig:pre-proccessing}
\end{figure}

The first step in using the extracted signals from intelligent tires is to find the acceleration signals that are related to the tire contact patch. In this section, we extract the data related to the tire contact patch in each tire revolution to be used for training the ML-based slip ratio estimation algorithm. Data preprocessing is mainly divided into filtering, identifying the tire contact patch, and data normalization. The whole process is shown in Fig. \ref{fig:pre-proccessing}.

\subsubsection{Filtering}
The data acquired by the DAQ system at 10 kHz are sampled for spectral analysis. Herein, 400 Hz is chosen as the cutoff frequency for this Butterworth low-pass filtering, as this frequency is mainly caused by tire deformation \cite{xu2020tire}. The higher frequency information can be used to identify road conditions \cite{lee2021intelligent} and tire slip angle \cite{xu2021tire}.

\subsubsection{Extracting the contact patch}

\begin{figure}[!h]
    \centering
    \includegraphics[width=3in]{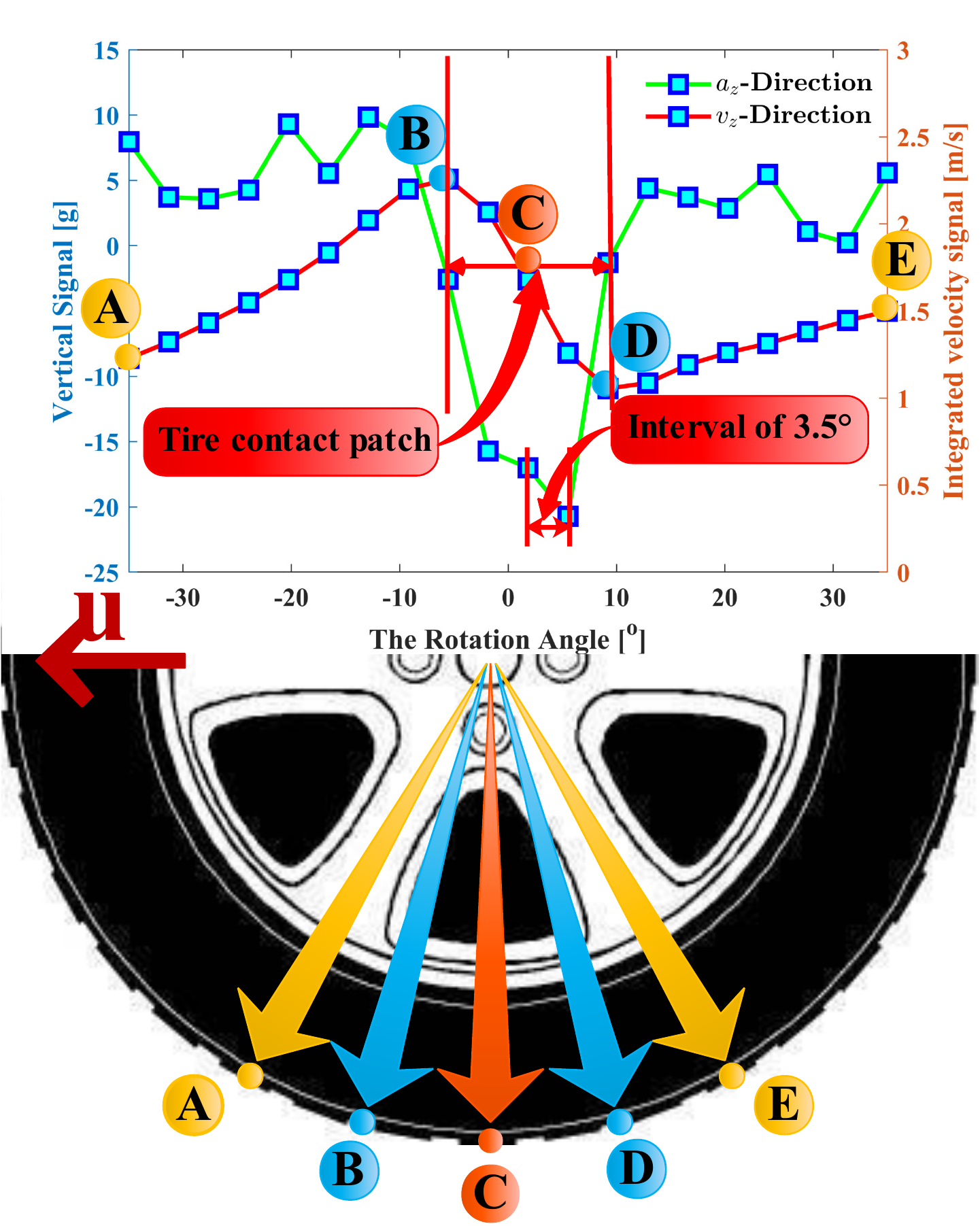}
\caption{Determining samples in the contact patch.}
    \label{fig:contact patch}
\end{figure}

Due to the instability of the longitudinal acceleration compared to vertical acceleration under longitudinal slip conditions, the method of determining the contact patch according to the leading and trailing peaks of the longitudinal acceleration is no longer applicable \cite{xu2020tire}. Thus, the integration of the vertical acceleration, i.e., the vertical velocity ($v_z$), is used to determine the contact patch (although the integration brings an offset, it does not affect the peak detection). As shown in Fig. \ref{fig:contact patch}, it can be clearly seen that there are two peaks in the vertical velocity when the sensor enters and leaves the contact patch. At the same time, the rotation angle of the tire is recorded by the encoder to accurately locate the position of the acceleration sensor. The two peaks of the vertical velocity (points $B$ and $D$, respectively) are observed to correspond to approximately $\pm{10^\circ}$ of the encoder, and the accelerations at $\pm{35^\circ}$ are extracted to provide more information about the contact patch for the ML algorithms (related to points $A$ and $E$). Additionally, the number of sampling points changes with increasing wheel angular velocity, so training samples can be captured according to the tire rotation angle. Meanwhile, to make the ML algorithm computationally efficient, a sample point is extracted at each $3.5^\circ$ interval \cite{xu2021tire}. This ensures that the acceleration information is extracted from the tire contact patch. Therefore, each tire rotation generates 20 sample points, which are used as inputs for training the ML algorithm.

\subsubsection{Data normalization}

To achieve fast convergence in the training of the ML algorithms used in this study, it is necessary to normalize the extracted data from the tire contact patch. Thus, the min-max normalization approach is used as follows:
\begin{align}
x_{norm} & = \frac{x-x_{min} }{x_{max} -x_{min}},
\label{eqn:02}
\end{align}
where $x$ presents the measured data and $x_{min}$ and $x_{max}$ are the minimum and maximum of the acquired data. It should be noted that the RF technique does not require the data normalization process.
\subsection{Machine Learning Algorithm Training}
In the area of tire and vehicle state estimations, the dominant ML techniques are decision tree algorithms such as decision trees, random forests \cite{xu2020tire} and gradient boosting machines \cite{xu2021tire}, and the the neural network family, including artificial neural networks \cite{lee2021intelligent}, convolutional networks, recurrent neural networks \cite{xu2020tire}; and support vector machines \cite{xu2021tire}. Considering the results of previous studies in this field, each of these techniques has its own advantages and disadvantages. Accordingly, we have used four different ML algorithms to provide a comprehensive study, which entails all ML families described above. We used an artificial neural network (ANN), gradient boosting machine (GBM), random forest (RF), and support vector machine (SVM) to estimate the tire slip ratio.

Generally, ANNs consist of input, hidden, and output layers. By increasing the number of hidden layers, we can construct a deeper model, which might lead to higher estimation accuracy. The implemented ANN works based on the resilient backpropagation (Rprop) algorithm, as it can effectively handle noise and is suitable for hardware-related applications \cite{askari2018towards,robert1989theory,askari2019towards}. A GBM is generally used for regression and classification purposes based on fusing weak and strong learners using iterative schemes \cite{friedman2001greedy}. Both RF and GBM are categorized as decision tree algorithms. Among these two techniques, the implementation of RF is easier, and it is less prone to overfitting \cite{ho1998random}. The fourth ML technique used in this paper is an SVM, which is suitable for both regression and classification analysis. In the case of solving linearly indistinguishable data, the kernel function technique is used to map vectors to a higher dimensional space. This results in establishing a maximal interval hyperplane (MIH) and building two parallel hyperplanes on both sides of the MIH that separate the data. The error of this technique depends on the distance of the abovementioned critical hyperplanes \cite{noble2006support}.

\begin{figure}[!h]
    \centering
    \includegraphics[width=3in]{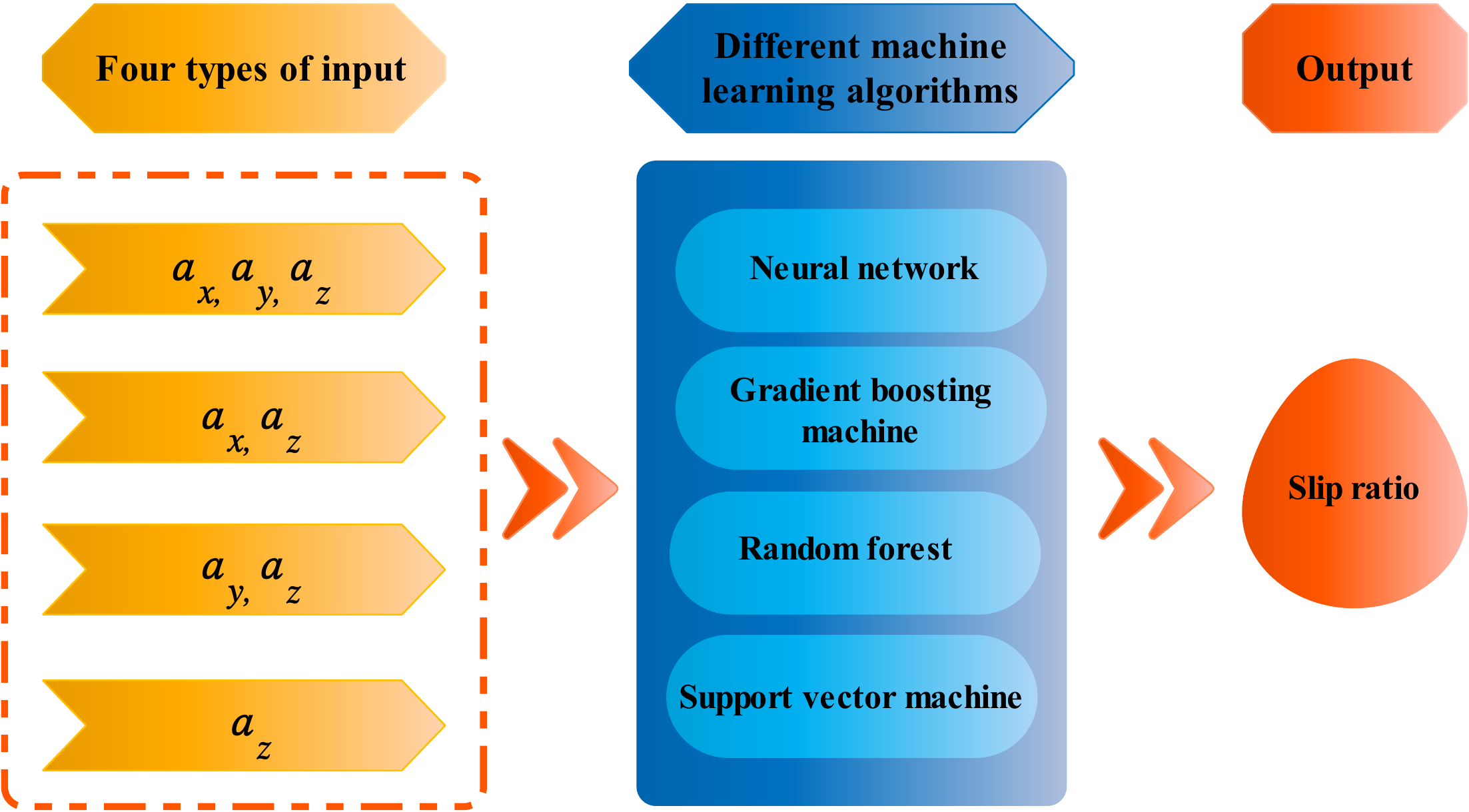}
\caption{Diagram of the inputs and output of different machine learning algorithms.}
    \label{fig:ML}
\end{figure}

Four different categories of inputs are used in this research for training the ML algorithms, including ($a_x,a_y,a_z$), ($a_x,a_z$), ($a_y,a_z$), and $a_z$. The output of all the ML algorithms is the slip ratio, as shown in Fig. \ref{fig:ML}. All training datasets in this research are conducted on a laptop with an Intel(R) Core(TM) i5-10200H CPU @ 2.4 GHz and 16 GB of DDR4 RAM. The hyperparameters used for all models are assigned by the "trial and error" method. The ANN is based on the Rprop algorithm for slip ratio estimation, and its hidden layer is 10-5-1. The hyperparameters of the GBM are as follows: trees = 4295, interaction.depth = 3, shrinkage = 0.1, n.minobinnode = 10, and bag.fraction = 1. The tree of the RF is chosen to be 50. The gamma of the SVM is set to 0.0312 with a cost of 32. Four categories of inputs are used with the same hyperparameters as above.

\section{Results and Discussion}
\label{sec:05}
This section is composed of two main parts. First, the results of slip ratio estimation based on various ML algorithms with four inputs are shown considering four working conditions. Second, to investigate the accuracy of the presented results, a 10-fold cross-validation (CV) method is used.

\subsection{Estimation Results of Machine Learning Algorithms}

We consider three aspects in the estimation of the tire slip ratio based on ML algorithms, including inputs, the type of ML algorithm, and the working condition of the data used for training. Based on the presented Data Analysis in Section \ref{sec:03}, we use $a_z$ as the basis for this study combined with the other two directions, i.e., ($a_x,a_y,a_z$), ($a_x,a_z$), ($a_y,a_z$), and $a_z$ as inputs to the ML algorithms used in this research. In addition, to verify the reliability and accuracy of each of the ML techniques used in this work, a validation study is carried out. We also studied the effect of working conditions on the performance of trained ML algorithms for slip ratio estimation.

\begin{table}[!h]
\caption{\MakeUppercase{Summary of test results based on different ML algorithms}}
\label{tab:my-table-02}
\begin{tabular}{cccccc}
\hline
\multicolumn{6}{c}{\textbf{Testing datasets NRMS errors (\%)}} \\ \hline
\multicolumn{1}{c|}{\begin{tabular}[c]{@{}c@{}}\textbf{Test} \\ \textbf{condition}\end{tabular}} & \multicolumn{1}{c|}{\begin{tabular}[c]{@{}c@{}}\textbf{Estimation} \\ \textbf{method}\end{tabular}} & \multicolumn{1}{c|}{\begin{tabular}[c]{@{}c@{}}$a_x$,$a_y$,\\ $a_z$\end{tabular}} & \multicolumn{1}{c|}{$a_x$,$a_z$} & \multicolumn{1}{c|}{$a_y$,$a_z$} & $a_z$ \\ \hline
\multicolumn{1}{c|}{\multirow{4}{*}{Data Set 1}} & \multicolumn{1}{c|}{Neural network} & \multicolumn{1}{c|}{2.62} & \multicolumn{1}{c|}{3.13} & \multicolumn{1}{c|}{2.77} & 2.17 \\ \cline{2-6} 
\multicolumn{1}{c|}{} & \multicolumn{1}{c|}{\begin{tabular}[c]{@{}c@{}}Gradient \\ boosting machine\end{tabular}} & \multicolumn{1}{c|}{4.60} & \multicolumn{1}{c|}{3.31} & \multicolumn{1}{c|}{3.98} & 3.02 \\ \cline{2-6} 
\multicolumn{1}{c|}{} & \multicolumn{1}{c|}{Random forest} & \multicolumn{1}{c|}{3.47} & \multicolumn{1}{c|}{2.60} & \multicolumn{1}{c|}{3.22} & 2.27 \\ \cline{2-6} 
\multicolumn{1}{c|}{} & \multicolumn{1}{c|}{\begin{tabular}[c]{@{}c@{}}Support \\ vector machine\end{tabular}} & \multicolumn{1}{c|}{4.83} & \multicolumn{1}{c|}{3.73} & \multicolumn{1}{c|}{3.87} & 3.04 \\ \hline
\multicolumn{1}{c|}{\multirow{4}{*}{Data Set 2}} & \multicolumn{1}{c|}{Neural network} & \multicolumn{1}{c|}{3.56} & \multicolumn{1}{c|}{2.78} & \multicolumn{1}{c|}{3.56} & 3.87 \\ \cline{2-6} 
\multicolumn{1}{c|}{} & \multicolumn{1}{c|}{\begin{tabular}[c]{@{}c@{}}Gradient \\ boosting machine\end{tabular}} & \multicolumn{1}{c|}{6.86} & \multicolumn{1}{c|}{5.62} & \multicolumn{1}{c|}{6.63} & 5.42 \\ \cline{2-6} 
\multicolumn{1}{c|}{} & \multicolumn{1}{c|}{Random forest} & \multicolumn{1}{c|}{6.75} & \multicolumn{1}{c|}{6.17} & \multicolumn{1}{c|}{7.58} & 5.41 \\ \cline{2-6} 
\multicolumn{1}{c|}{} & \multicolumn{1}{c|}{\begin{tabular}[c]{@{}c@{}}Support \\ vector machine\end{tabular}} & \multicolumn{1}{c|}{5.46} & \multicolumn{1}{c|}{4.50} & \multicolumn{1}{c|}{4.48} & 3.18 \\ \hline
\multicolumn{1}{c|}{\multirow{4}{*}{Data Set 3}} & \multicolumn{1}{c|}{Neural network} & \multicolumn{1}{c|}{11.08} & \multicolumn{1}{c|}{10.16} & \multicolumn{1}{c|}{9.59} & 9.27 \\ \cline{2-6} 
\multicolumn{1}{c|}{} & \multicolumn{1}{c|}{\begin{tabular}[c]{@{}c@{}}Gradient \\ boosting machine\end{tabular}} & \multicolumn{1}{c|}{14.06} & \multicolumn{1}{c|}{12.58} & \multicolumn{1}{c|}{15.21} & 12.76 \\ \cline{2-6} 
\multicolumn{1}{c|}{} & \multicolumn{1}{c|}{Random forest} & \multicolumn{1}{c|}{15.37} & \multicolumn{1}{c|}{14.25} & \multicolumn{1}{c|}{16.64} & 13.41 \\ \cline{2-6} 
\multicolumn{1}{c|}{} & \multicolumn{1}{c|}{\begin{tabular}[c]{@{}c@{}}Support \\ vector machine\end{tabular}} & \multicolumn{1}{c|}{19.61} & \multicolumn{1}{c|}{17.02} & \multicolumn{1}{c|}{16.27} & 12.95 \\ \hline
\multicolumn{1}{c|}{\multirow{4}{*}{Data Set 4}} & \multicolumn{1}{c|}{Neural network} & \multicolumn{1}{c|}{10.21} & \multicolumn{1}{c|}{5.61} & \multicolumn{1}{c|}{7.64} & 6.01 \\ \cline{2-6} 
\multicolumn{1}{c|}{} & \multicolumn{1}{c|}{\begin{tabular}[c]{@{}c@{}}Gradient \\ boosting machine\end{tabular}} & \multicolumn{1}{c|}{6.30} & \multicolumn{1}{c|}{5.36} & \multicolumn{1}{c|}{6.39} & 5.58 \\ \cline{2-6} 
\multicolumn{1}{c|}{} & \multicolumn{1}{c|}{Random forest} & \multicolumn{1}{c|}{7.81} & \multicolumn{1}{c|}{7.19} & \multicolumn{1}{c|}{7.45} & 6.89 \\ \cline{2-6} 
\multicolumn{1}{c|}{} & \multicolumn{1}{c|}{\begin{tabular}[c]{@{}c@{}}Support \\ vector machine\end{tabular}} & \multicolumn{1}{c|}{12.36} & \multicolumn{1}{c|}{10.76} & \multicolumn{1}{c|}{9.32} & 7.30 \\ \hline
\end{tabular}
\end{table}

\begin{figure}[!h]
    \centering
    \includegraphics[width=3in]{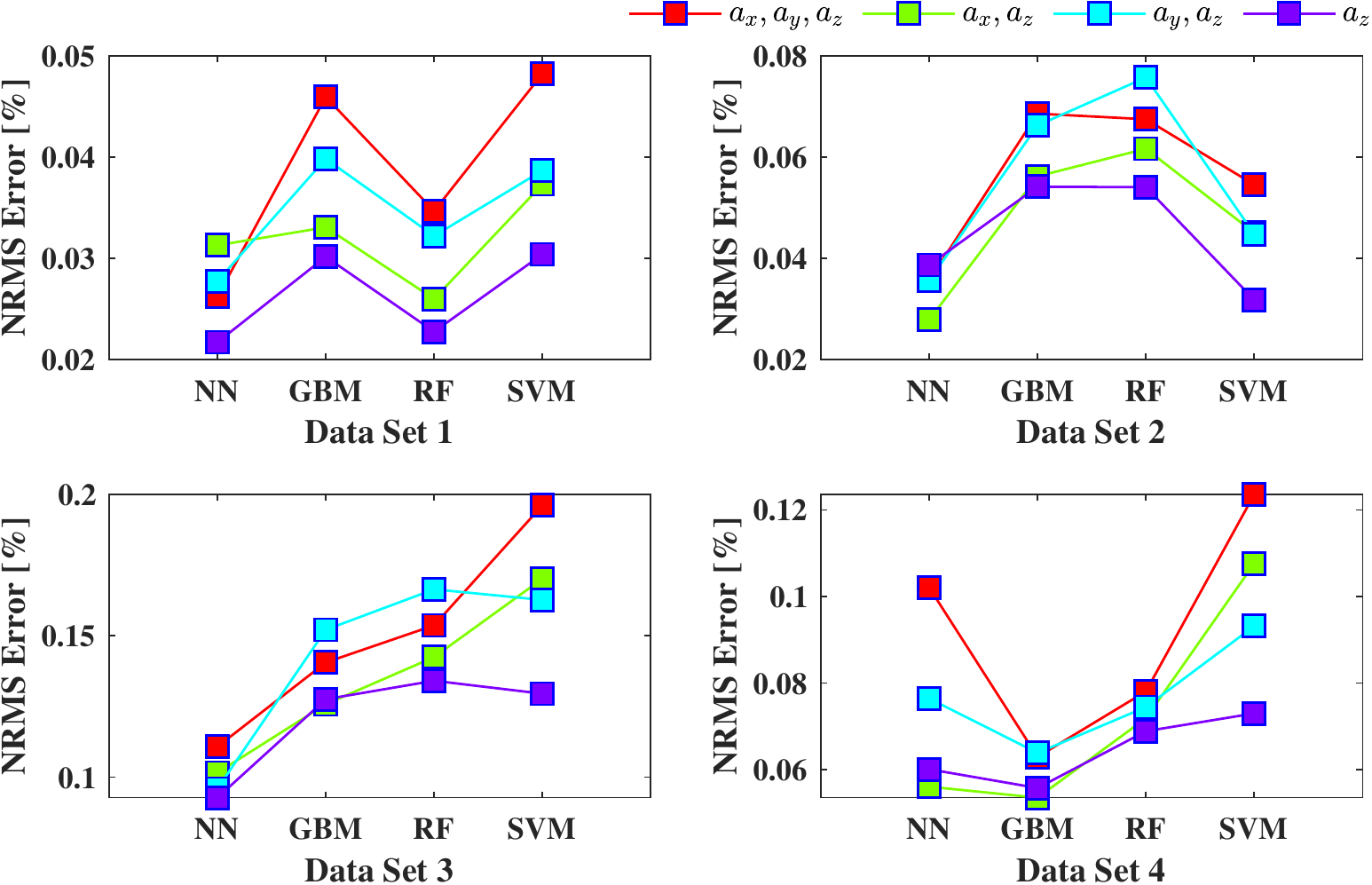}
\caption{Graphical representation of the NRMS errors for datasets used: (a) Data Set 1; (b) Data Set 2; (c) Data Set 3; (d) Data Set 4.}
    \label{fig:All_inf}
\end{figure}

Table \ref{tab:my-table-02} shows the normalized root mean square (NRMS) error of the slip ratio considering the abovementioned influencing factors. The effect of the first factor (input categories) is also depicted in Fig. \ref{fig:All_inf}. As shown, the slip ratio estimation based on $a_z$ data has the highest accuracy regardless of the class of dataset and type of ML algorithms. Second, in terms of the type of ML algorithm, the ANN performs well except for Data Set 4 with ($a_x,a_y,a_z$) as input. The next best performer is GBM, especially in Data Set 4, where the estimation results are stable for all four inputs. In the final step, a study is performed based on the estimation results of all datasets. The NRMS errors are relatively small for Data Set 1 and Data Set 2, as they are mainly related to small slip ratio conditions. The results obtained based on the RF algorithm using ($a_y,a_z$) as input in Data Set 2 have the highest error in comparison with other techniques. The NRMS error is increased up to 15\% as the slip ratio increases based on Data Set 3.

\begin{figure}[!h]
    \centering
    \includegraphics[width=3in]{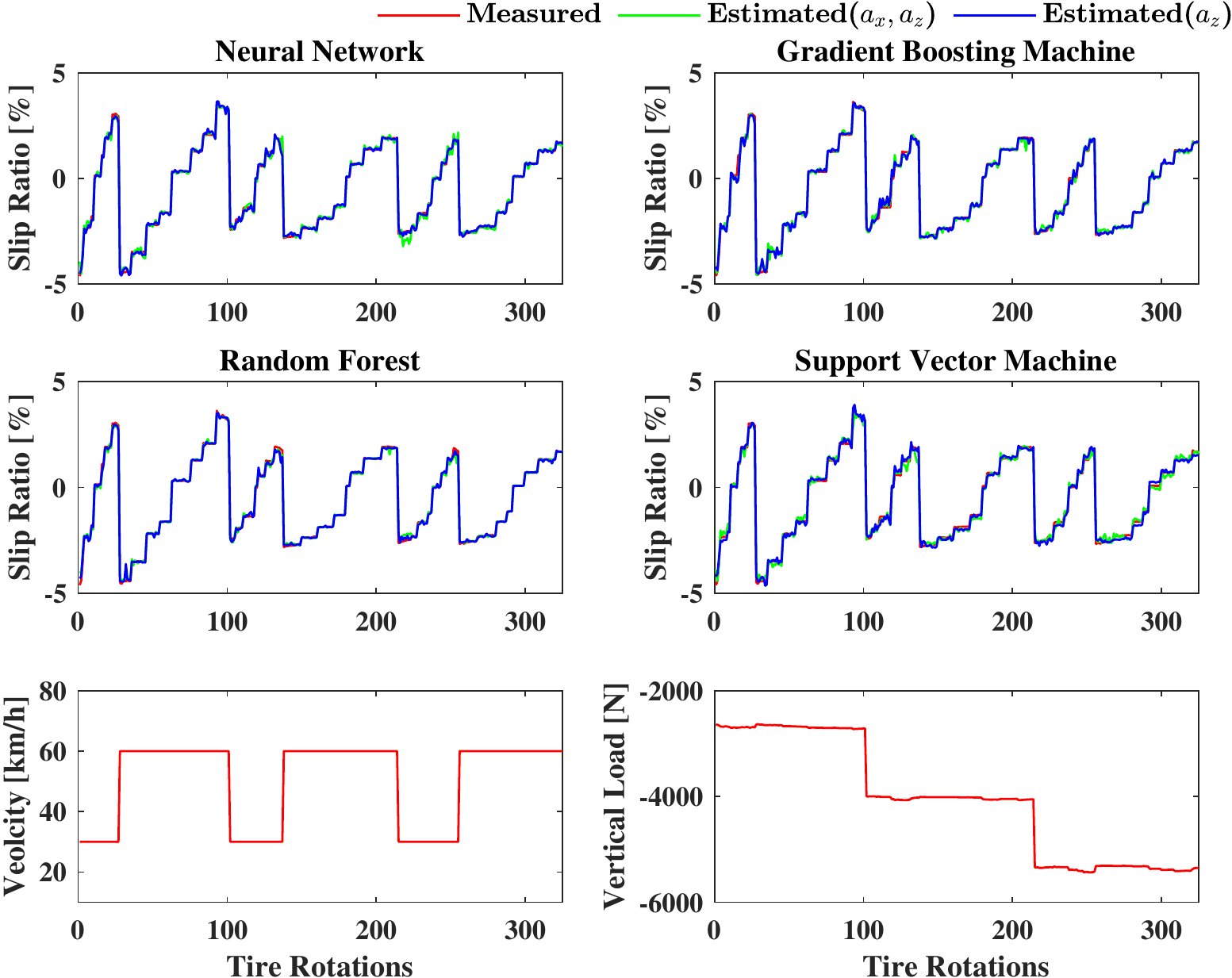}
\caption{Comparison of estimated slip ratio with different ML methods in Data Set 1.}
    \label{fig:SSR}
\end{figure}

\begin{figure}[!h]
    \centering
    \includegraphics[width=3in]{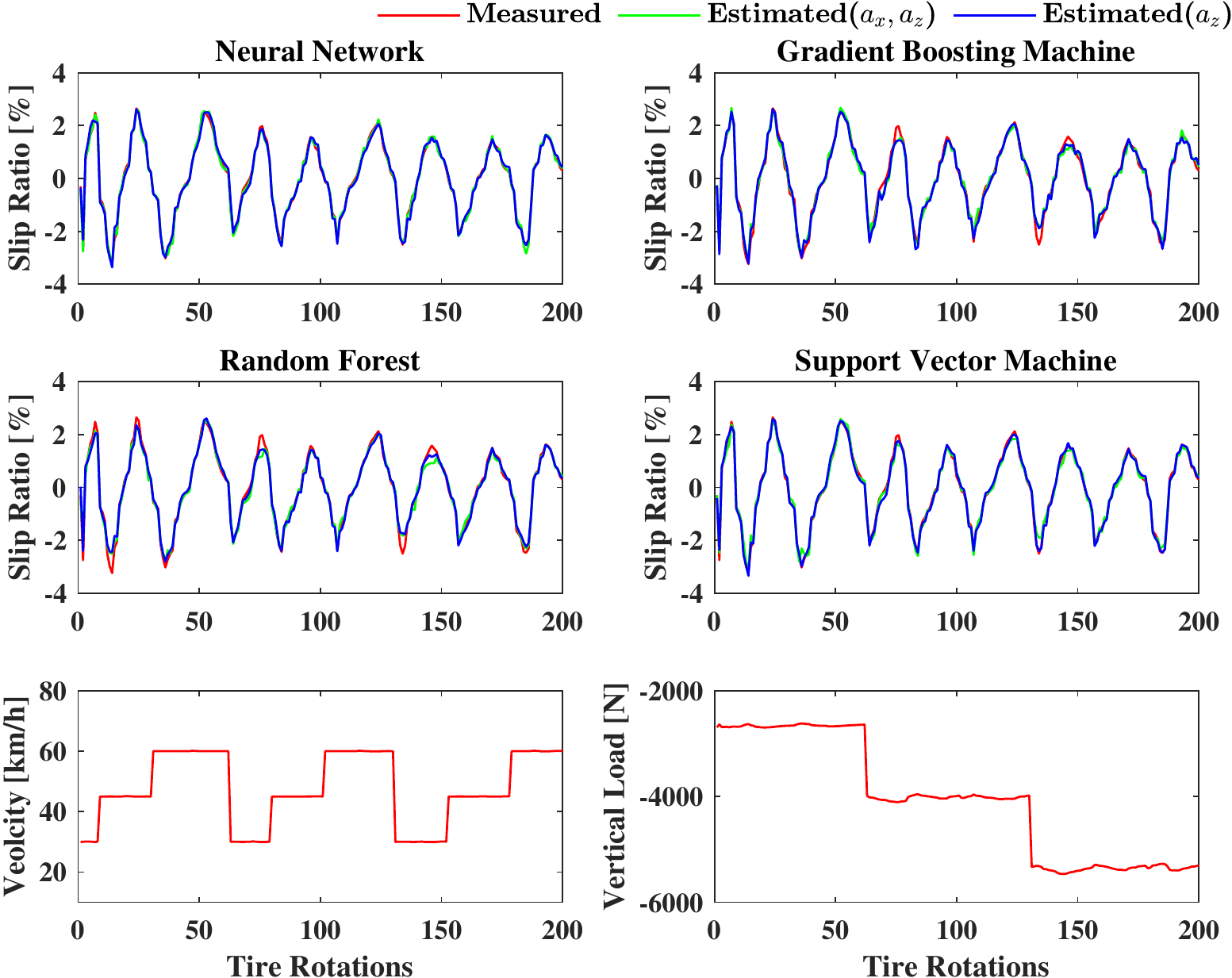}
\caption{Comparison of estimated slip ratio with different ML methods in Data Set 2.}
    \label{fig:SCSR}
\end{figure}

\begin{figure}[!h]
    \centering
    \includegraphics[width=3in]{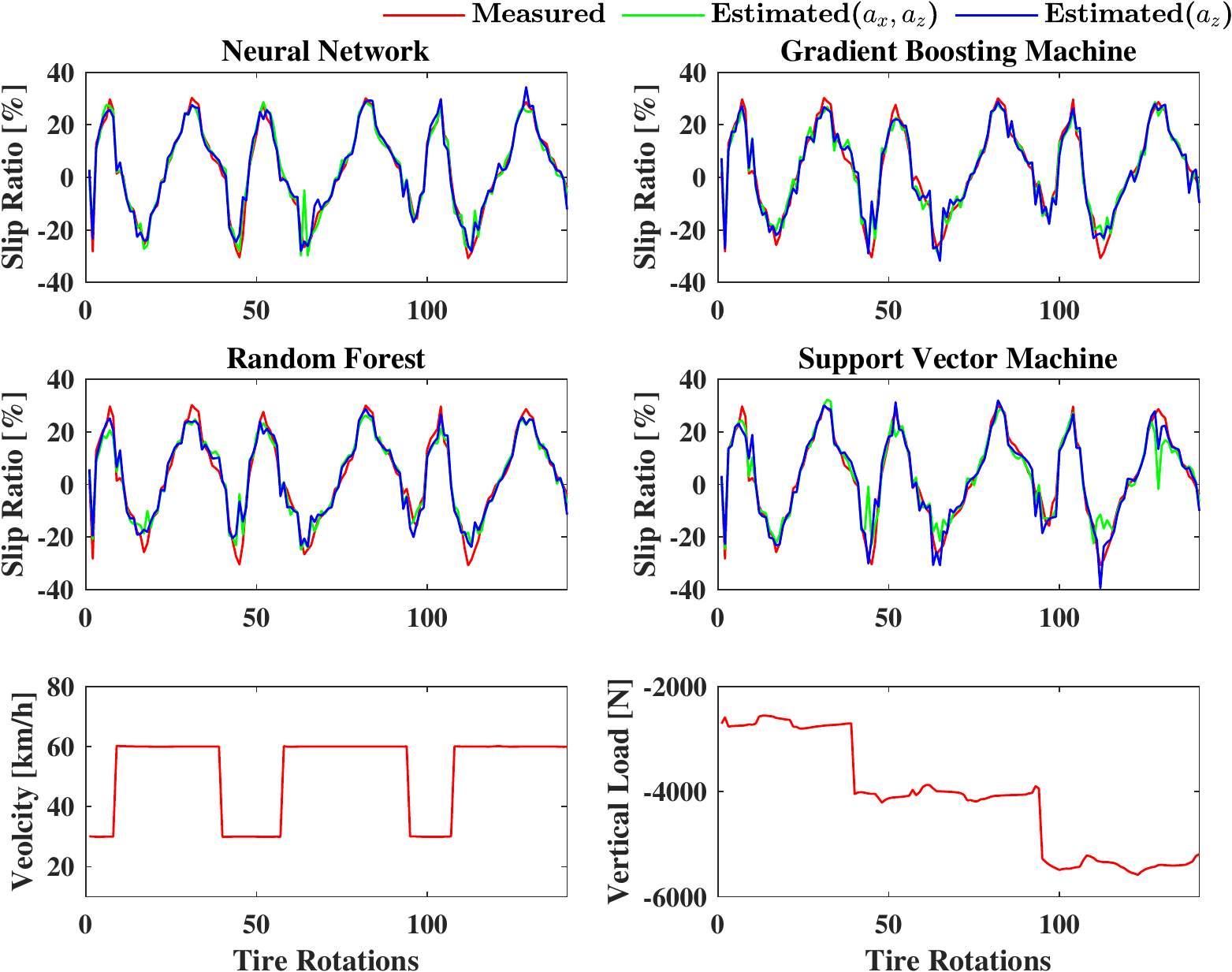}
\caption{Comparison of estimated slip ratio with different ML methods in Data Set 3.}
    \label{fig:LCSR}
\end{figure}

\begin{figure}[!h]
    \centering
    \includegraphics[width=3in]{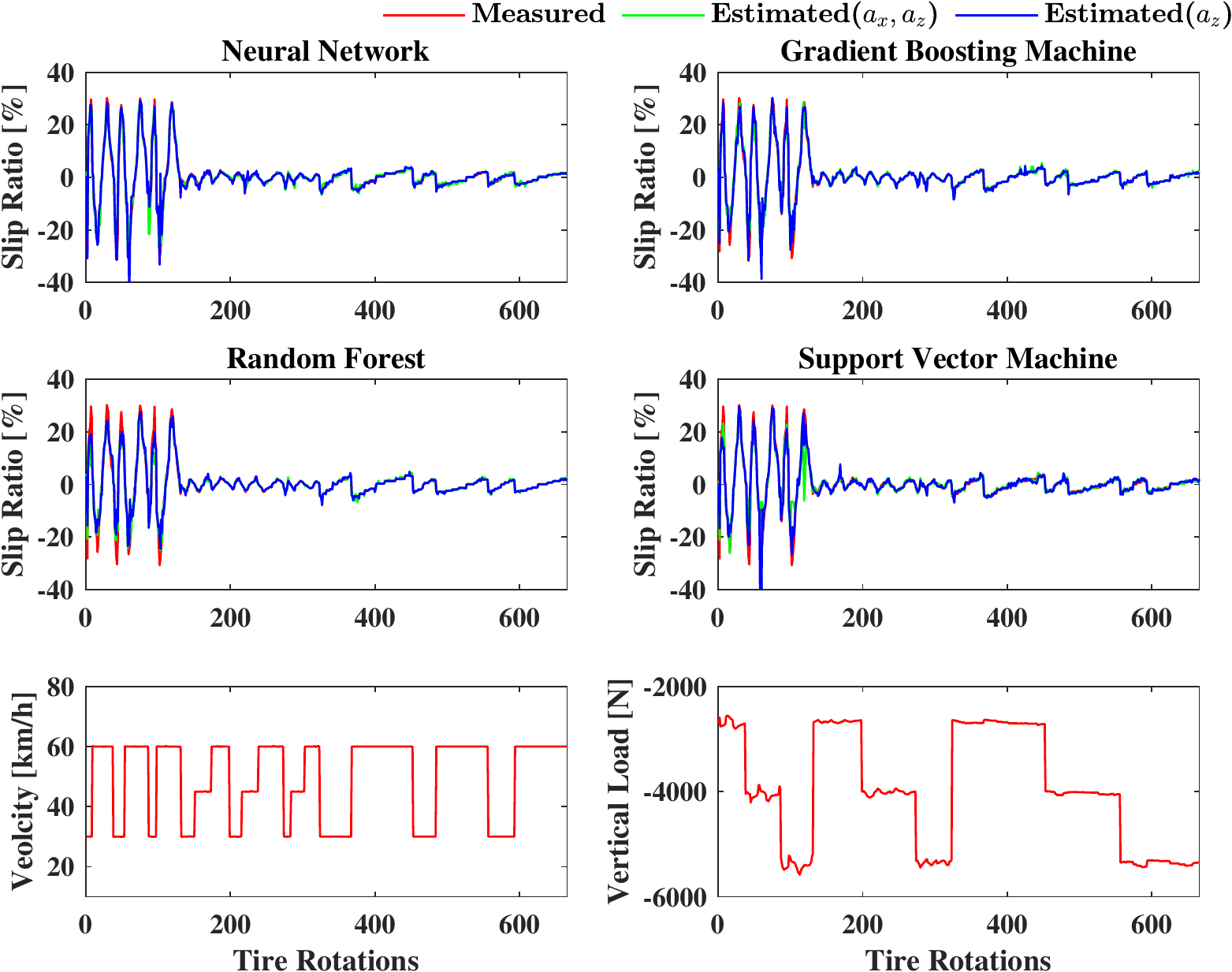}
\caption{Comparison of estimated slip ratio with different ML methods in Data Set 4.}
    \label{fig:AllCondi}
\end{figure}

In addition to the statistical results described above, two types of algorithm inputs, ($a_x,a_z$) and $a_z$, are selected for a more detailed evaluation to graphically compare the estimated slip ratio with the measured slip ratio. Figs. \ref{fig:SSR}-\ref{fig:AllCondi} provide a comparison between the estimated results and measured values considering all data sets. Each subplot represents the results obtained by an ML algorithm utilized in this research. As Fig. \ref{fig:SSR} shows, ML algorithms estimate the tire slip ratio with very good accuracy for Data Set 1 irrespective of velocity and load. Similar behavior is observed in Fig. \ref{fig:SCSR} in terms of estimation accuracy except in the GBM and RF results, where tires experience a relatively large slip ratio (Data Set 2). The tabulated results in Table \ref{tab:my-table-02} also reveal similar results in terms of NRMS errors for GBM and RF algorithms. As Table II shows, the NRMS errors are 5.62\% and 6.17\% for GBM and RF with the use ($a_x,a_z$) as the input. When $a_z$ is used as the input, the error is reduced to 5.42\% and 5.41\% for the GBM and RF algorithms, respectively. As Fig. \ref{fig:LCSR} shows, Data Set 3 entails the highest slip ratio of 30\% in which the maximum estimation error occurs (around the slip ratio of 30\%). This conclusion is identical regardless of the ML algorithms and inputs. In addition, the ANN shows the highest accuracy even when the tire experiences a high slip ratio. Fig. \ref{fig:AllCondi} shows the estimation results of Data Set 4, and it can be seen that the ANN algorithm provides the best results in terms of tire slip ratio estimation compared to other techniques.

\subsection{10-Fold Cross-Validation}

To effectively test the performance of different ML algorithms on unknown data, $k$-fold cross-validation (CV) is used in this section of the paper. The training dataset is split into $k$ subsamples in which one of them is used as testing data and the remaining ($k$-1) subsamples are used for training purposes. $k$-fold CV is repeated so that each subsample can be sufficiently utilized.

\begin{figure}[!h]
    \centering
    \includegraphics[width=3in]{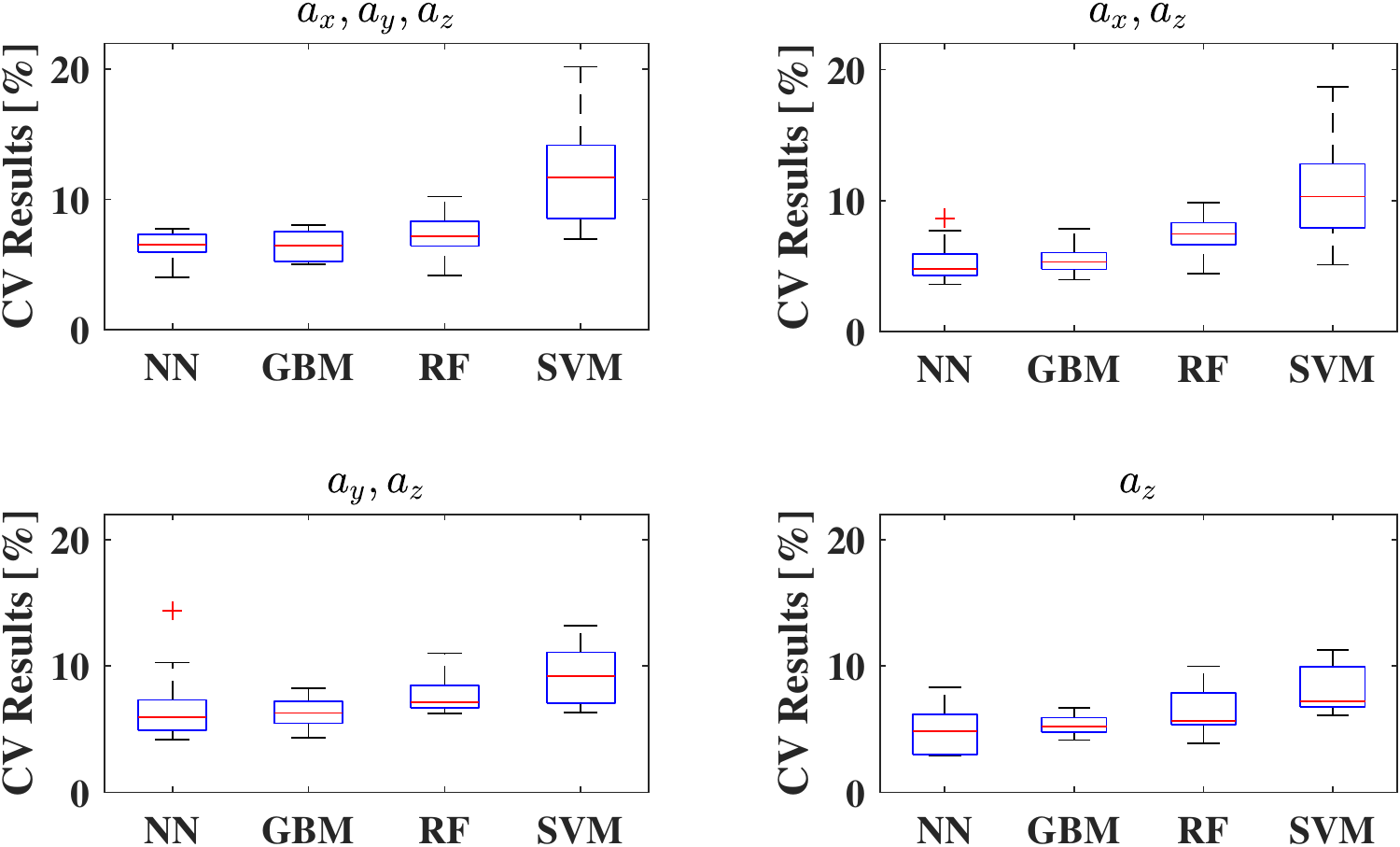}
\caption{Boxplot of CV results for different algorithm inputs: (a) ($a_x,a_y,a_z$) as input; (b) ($a_x,a_z$) as input; (c) ($a_y,a_z$) as input; (d) $a_z$ as input.}
    \label{fig:CV-01}
\end{figure}

\begin{figure}[!h]
    \centering
    \includegraphics[width=3in]{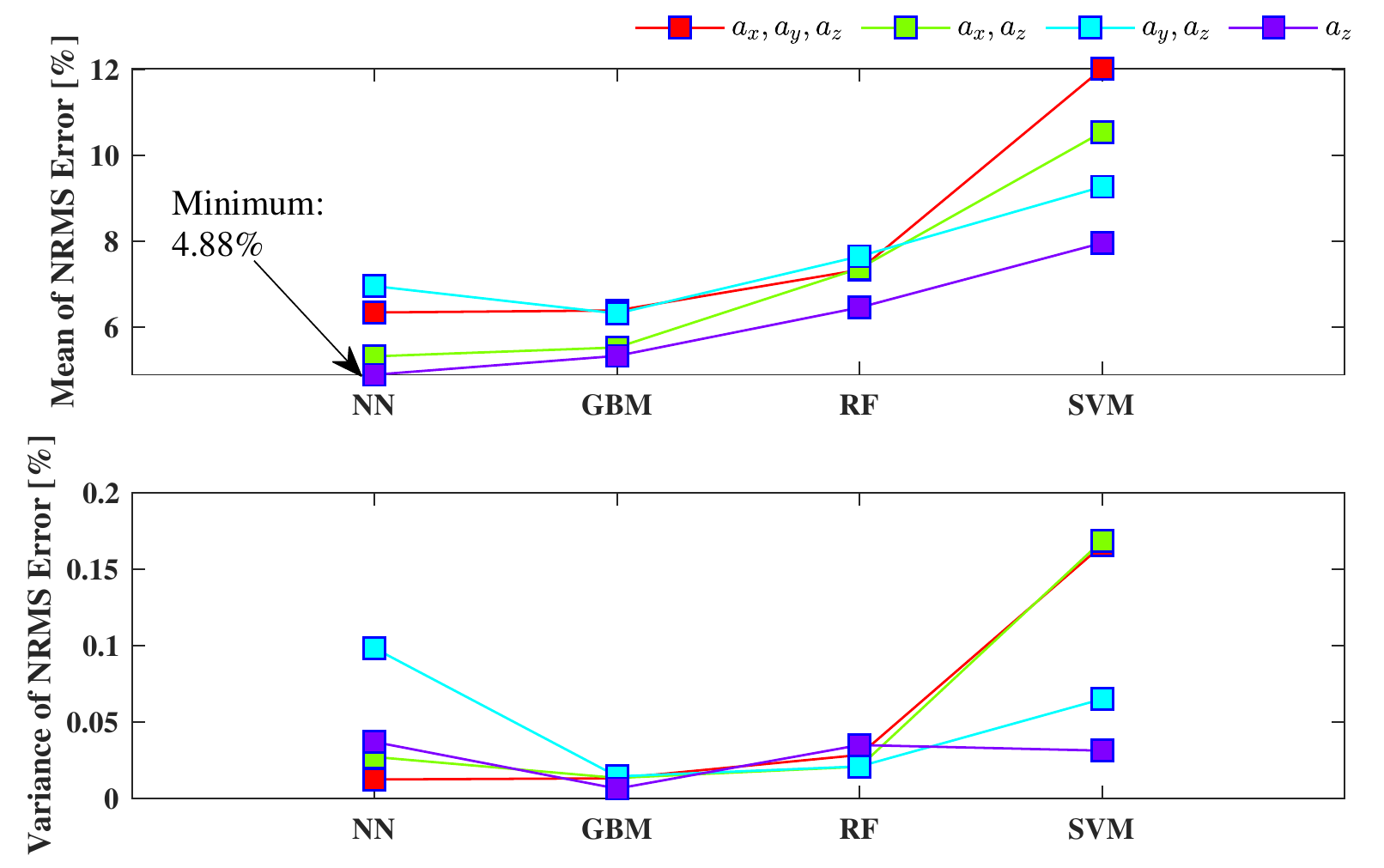}
\caption{Comparison of mean and variance for CV results.}
    \label{fig:CV-02}
\end{figure}

Data Set 4 contains all the datasets for this experiment, which is used to perform the 10-fold cross-validation. Fig. \ref{fig:CV-01} shows the boxplot for the cross-validation of different machine learning methods for the four algorithm inputs. The mean and variance of the cross-validation results are plotted in Fig. \ref{fig:CV-02}. As Fig. \ref{fig:CV-02} shows, the ANN has the smallest NRMS error average with the $a_z$ input (4.88\%). Considering the algorithm types and combining the results in Figs. \ref{fig:CV-01} and \ref{fig:CV-02}, the GBM has an almost similar mean value to the ANN, with a smaller NRMS error variance. However, the ANN has the lowest NRMS error, and the GBM has the best performance considering its lower error variance. In particular, outliers appear in the cross-validation results for each of ($a_x,a_z$) and ($a_y,a_z$) as inputs in Fig. \ref{fig:CV-01}. In addition, the $a_z$ input model performs better overall, with the smallest mean and relatively small variance. This means that if only $a_z$ is used as input, it allows the ML-based slip ratio estimation algorithm to run with fewer datasets and reduces computational costs, which offers the possibility of real-time operation of real vehicles in the future.

\section{Conclusion}
\label{sec:06}

In this article, the tire slip ratio was estimated using an intelligent tire system and machine learning algorithms. An intelligent tire system was developed with the use of an accelerometer attached to the inner liner of the tire. The three-directional acceleration signals generated by the accelerometers were used for a systematic and exhaustive analysis. Promising features were identified, especially the vertical acceleration, and then used to train four different machine learning algorithms, including an ANN, GBM, RF, and SVM, to estimate the tire slip ratio under different working conditions. The estimation results indicate that ML techniques have promising potential in the accurate estimation of the tire slip ratio. To further evaluate the potency and reliability of the implemented ML algorithms, a 10-fold CV method was used. It was shown that the mean value of GBM NRMS error is relatively small with the minimum variance compared to other ML techniques. The ANN is similar to the GBM but has relatively poor stability. It was also revealed that the trained ML in which $a_z$ signals have been used as the input provides higher accuracy in the estimation of the tire slip ratio. The results obtained based on this research show the high potential of intelligent tire systems and machine learning algorithms for tire slip ratio estimation. For future research, we will extend more training samples (for different tire brands, tire pressures, and even combined working conditions) to improve the performance of the intelligent tire system for slip ratio estimation, and also develop research on tire slip angle, tire longitudinal force, and tire lateral force estimation.

\bibliographystyle{IEEEtran}
\bibliography{myref}

\begin{IEEEbiography}[{\includegraphics[width=1in,height=1.25in,clip,keepaspectratio]{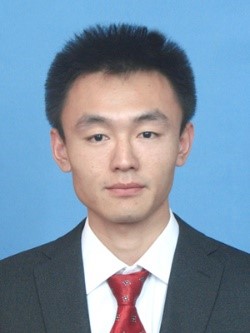}}]{Nan Xu} received a Ph.D. degree from the State Key Laboratory of Automotive Simulation and Control, Jilin University, Changchun, China, in 2012. He is currently an associate professor with the State Key Laboratory of Automotive Simulation and Control, Jilin University. He was a Visiting Scholar with the Department of Mechanical and Mechatronics Engineering, University of Waterloo. His current research focuses on tire dynamics, intelligent tires, dynamics and stability control of electric vehicles and autonomous vehicles.
\end{IEEEbiography}

\begin{IEEEbiography}[{\includegraphics[width=1in,height=1.25in,clip,keepaspectratio]{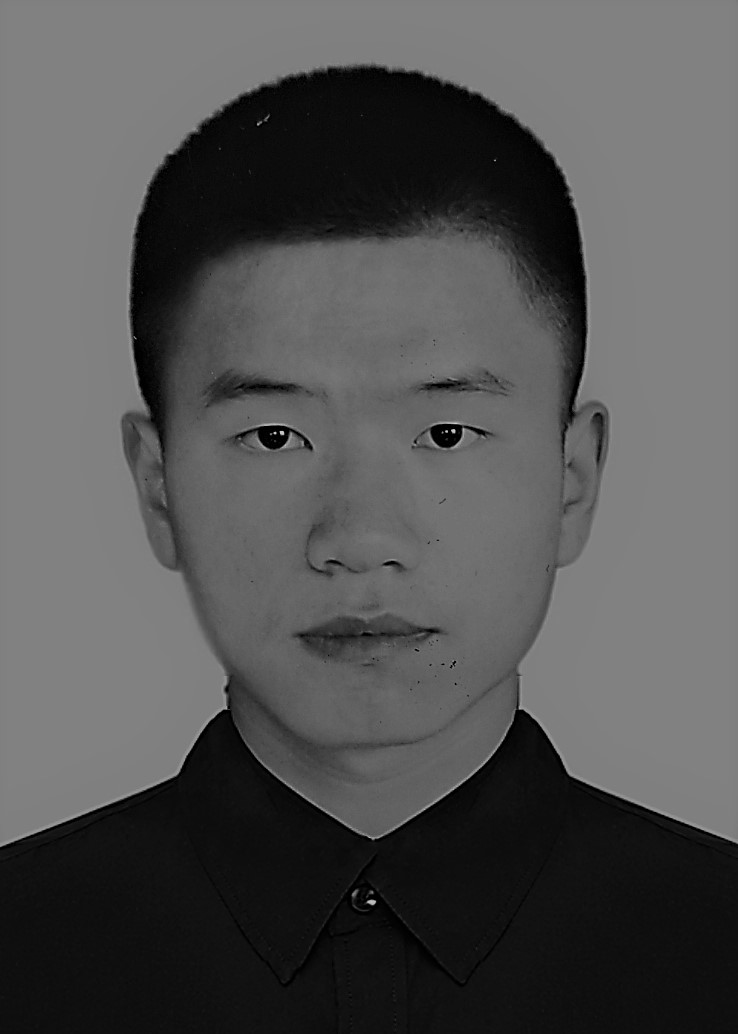}}]{Zepeng Tang} was born in Tongguan, Shaanxi, China in 1995. He received a B.E. degree from Tianjin University of Science and Technology. He is currently a M.S. candidate at the College of Automotive Engineering, Jilin University, Changchun, China. His current research interest focuses on intelligent tires, vehicle dynamics and autonomous vehicles.
\end{IEEEbiography}

\begin{IEEEbiography}[{\includegraphics[width=1in,height=1.25in,clip,keepaspectratio]{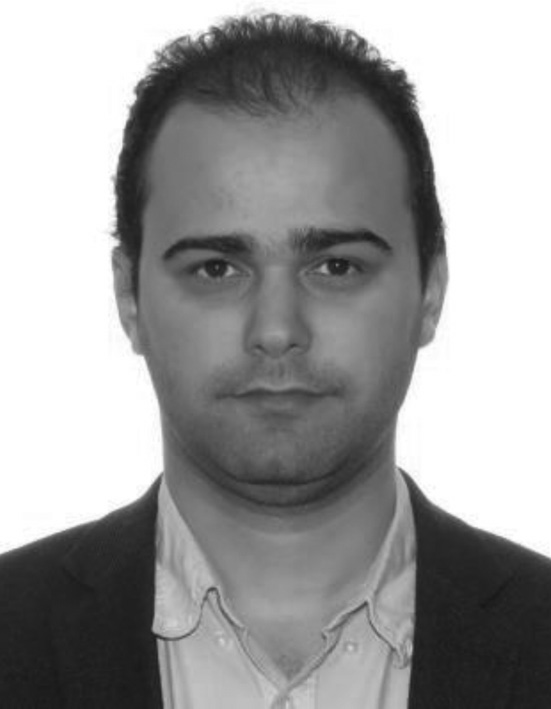}}]{Hassan Askari} was born in Rasht, Iran and received his B. Sc. ,M.Sc. and PhD degrees from Iran University of Science and Technology, Tehran, Iran, University of Ontario Institute of Technology, Oshawa, Canada, and University of Waterloo, Waterloo, Canada in 2011, 2014, and 2019, respectively. He has published more than 70 journal and conference papers in the areas of nonlinear vibrations, applied mathematics, nanogenerators and self-powered sensors. He coauthored one book and one book chapter published by Springer. He is an active reviewer for more than 40 journals and editorial board members of several scientific and international journals. He has received several prestigious awards, including Outstanding Researcher at the Iran University of Science and Technology, Fellowship of the Waterloo Institute of Nanotechnology, NSERC Graduate Scholarship, Ontario Graduate Scholarship, and the University of Waterloo President Award. He was nominated for the Governor General's Academic Gold Medal at the University of Ontario Institute of Technology and University of Waterloo in 2014 and 2019, respectively. He is currently a Postdoctoral Fellow at the Department of Mechanical and Mechatronics Engineering at the University of Waterloo.
\end{IEEEbiography}

\begin{IEEEbiography}[{\includegraphics[width=1in,height=1.25in,clip,keepaspectratio]{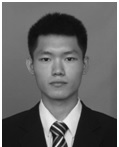}}]{Jianfeng Zhou} received a B.E. degree in automotive engineering in 2018 from Jinlin University, Changchun, China, where he is currently working toward a Ph.D. degree. His current research focuses on tire dynamics, intelligent tires and vehicle dynamics.
\end{IEEEbiography}

\begin{IEEEbiography}[{\includegraphics[width=1in,height=1.25in,clip,keepaspectratio]{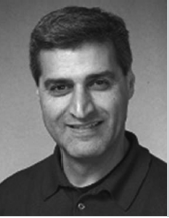}}]{Amir Khajepour} is currently a Professor of mechanical and mechatronics engineering at the University of Waterloo, Waterloo, ON, Canada, where he is also the Canada Research Chair in mechatronic vehicle systems. He has developed an extensive research program that applies his expertise in several key multidisciplinary areas. He is a fellow of The Engineering Institute of Canada, The American Society of Mechanical Engineers, and The Canadian Society of Mechanical Engineering.
\end{IEEEbiography}

\end{document}